%% file: main.tex
\documentclass[10pt,twocolumn]{article}

\usepackage{microtype}
\usepackage{graphicx}

\usepackage{cuted}
\usepackage{booktabs} 
\usepackage{multirow} 
\usepackage{dashrule}
\usepackage{arydshln} 
\usepackage{tabularray}

\usepackage{tikz}
\usetikzlibrary{arrows,intersections}
\tikzset{font=\scriptsize}
\usepackage{pgfplots}
\pgfplotsset{compat=1.11}
\usepgfplotslibrary{fillbetween}
\usetikzlibrary{backgrounds}

\definecolor{plotsgreen}{RGB}{38,150,38}
\definecolor{plotsorange}{RGB}{255,115,17}
\definecolor{plotsred}{RGB}{208,34,35}
\definecolor{plotsyellow}{RGB}{255,225,80}
\definecolor{plotsgrey}{RGB}{128,128,128}
\definecolor{plotspurple}{RGB}{137,92,181}
\definecolor{plotsblue}{RGB}{0,101,167}

\usepackage{subcaption} 
\usepackage{array}

\newcommand{\modelname}{DeepFlow\xspace}
\newcommand{\blockname}{VeRA\xspace}

\usepackage{xcolor}
\definecolor{blue}{rgb}{0.22, 0.22, 0.95}
\definecolor{bblue}{rgb}{0.12, 0.43, 0.84}
\definecolor{gray}{rgb}{0.29, 0.31, 0.31}
\definecolor{green}{rgb}{0.22, 0.7, 0.22}
\definecolor{lred}{rgb}{0.85, 0.27, 0.08}
\definecolor{tred}{rgb}{0.459,0.184,0.063}
\definecolor{orange}{rgb}{1.0, 0.4, 0}

\newcommand{\app}{\raise.17ex\hbox{$\scriptstyle\sim$}}

\newcolumntype{x}[1]{>{\centering\arraybackslash}p{#1pt}}

\newlength\savewidth\newcommand\shline{\noalign{\global\savewidth\arrayrulewidth
  \global\arrayrulewidth 1pt}\hline\noalign{\global\arrayrulewidth\savewidth}}
\newcommand{\tablestyle}[2]{\setlength{\tabcolsep}{#1}\renewcommand{\arraystretch}{#2}\centering\footnotesize}
\makeatletter\renewcommand\paragraph{\@startsection{paragraph}{4}{\z@}
  {.5em \@plus1ex \@minus.2ex}{-.5em}{\normalfont\normalsize\bfseries}}\makeatother

\usepackage{iccv}
\usepackage{float}
\usepackage{wrapfig}

\usepackage{colortbl}

\definecolor{baselinecolor}{gray}{.9}
\definecolor{darkbluecolor}{rgb}{0.63, 0.72, 0.81}
\definecolor{graybluecolor}{rgb}{0.68, 0.74, 0.77}
\definecolor{lightbluecolor}{rgb}{0.73, 0.83, 0.93}
\definecolor{lighterbluecolor}{rgb}{0.86, 0.91, 0.96}

\definecolor{darkredcolor}{rgb}{0.95, 0.71, 0.60}
\definecolor{lightredcolor}{rgb}{0.96, 0.82, 0.75}
\definecolor{lighterredcolor}{rgb}{0.97, 0.90, 0.87}

\newcommand{\baseline}[1]{\cellcolor{baselinecolor}{#1}}
\newcommand{\darkblue}[1]{\cellcolor{darkbluecolor}{#1}}
\newcommand{\darkred}[1]{\cellcolor{darkredcolor}{#1}}

\newcommand{\lightred}[1]{\cellcolor{lightredcolor}{#1}}
\newcommand{\lighterred}[1]{\cellcolor{lighterredcolor}{#1}}
\newcommand{\lightblue}[1]{\cellcolor{lightbluecolor}{#1}}
\newcommand{\lighterblue}[1]{\cellcolor{lighterbluecolor}{#1}}
\definecolor{darkergreen}{RGB}{10,100,10}

\usepackage{amsmath}
\usepackage{amssymb}
\usepackage{mathtools}
\usepackage{amsthm}

\usepackage{pifont} 

\theoremstyle{plain}

\theoremstyle{definition}

\theoremstyle{remark}

\definecolor{citecolor}{RGB}{34,139,34}
\usepackage[pagebackref=true,breaklinks=true,colorlinks,
  citecolor=citecolor,bookmarks=false]{hyperref}

\usepackage[capitalize,noabbrev]{cleveref}

\def\iccvPaperID{***}\iccvfinalcopy 

\begin{document}

\title{Deeply Supervised Flow-Based Generative Models}

\author{Inkyu Shin \quad Chenglin Yang \quad  Liang-Chieh Chen \vspace{3mm}\\
ByteDance Seed \vspace{1.5mm}\\
\href {https://deepflow-project.github.io/} {https://deepflow-project.github.io} 
}

\maketitle


\input{sections/0.abstract}
\input{sections/1.introduction}
\input{sections/2.related_works}

\input{sections/3.method}

\input{sections/4.experiments}
\input{sections/5.conclusion}

\clearpage

\input{sections/6.appendix}

\clearpage
\newpage
\clearpage
{\small
\bibliographystyle{ieee_fullname}
\bibliography{reference}
}





\end{document}

%% file: sections/0.abstract.tex
\begin{abstract}

Flow-based generative models have charted an impressive path across multiple visual generation tasks by adhering to a simple principle: learning velocity representations of a linear interpolant. However, we observe that training velocity solely from the final layer's output 
under-utilizes the rich inter-layer representations, potentially impeding model convergence. To address this limitation, we introduce \textbf{\modelname}, a novel framework that enhances velocity representation through inter-layer communication. 
\modelname partitions transformer layers into balanced branches with deep supervision and inserts a lightweight Velocity Refiner with Acceleration (\blockname) block between adjacent branches, which aligns the intermediate velocity features within transformer blocks.
Powered by the improved deep supervision via the internal velocity alignment, \modelname converges \textbf{8× faster} on ImageNet-256×256 with equivalent performance and further reduces FID by \textbf{2.6} while halving training time compared to previous flow-based models without a classifier-free guidance.
\modelname also outperforms baselines in text-to-image generation tasks, as evidenced by evaluations on MS-COCO and zero-shot GenEval.

\input{figures/efficiency_fid}
\input{figures/intro_featdist_fid}
\end{abstract}

%% file: figures/efficiency_fid.tex
\begin{figure}[t!]
\begin{center}
\includegraphics[width=0.99\linewidth]{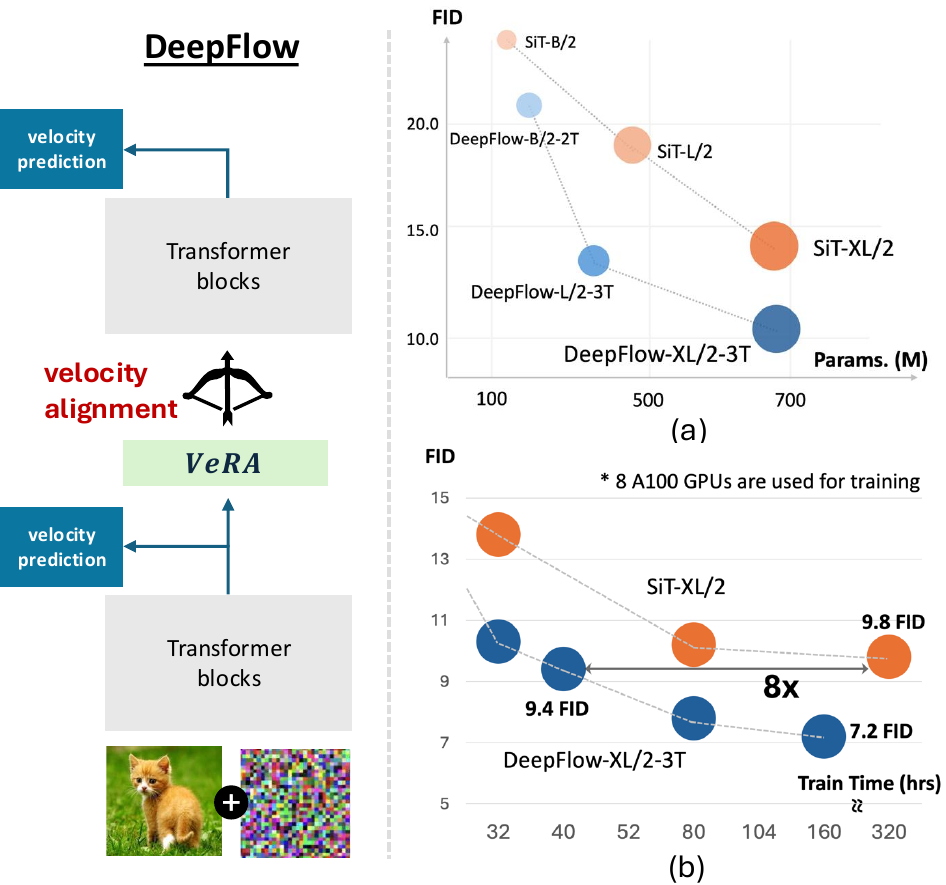}
\caption{
\textbf{Overview of \modelname.}
\textit{Left:} \modelname incorporates deep supervision by evenly adding velocity prediction within transformer blocks, further enhanced by the proposed Velocity Alignment block (\blockname).
\textit{Right:} (a) On the ImageNet-256 benchmark, \modelname consistently outperforms SiT~\cite{ma2024sit} in FID scores across various model sizes. (b) \modelname-XL achieves an 8$\times$ training efficiency improvement over SiT-XL. See~\cref{tab:param} for details.
}
\label{fig:eff_fid}
\end{center}
\vspace{-4mm}
\end{figure}
\vspace{-4mm}

%% file: figures/intro_featdist_fid.tex
\begin{figure*}[t]
\begin{center}
\includegraphics[width=1.0\linewidth]{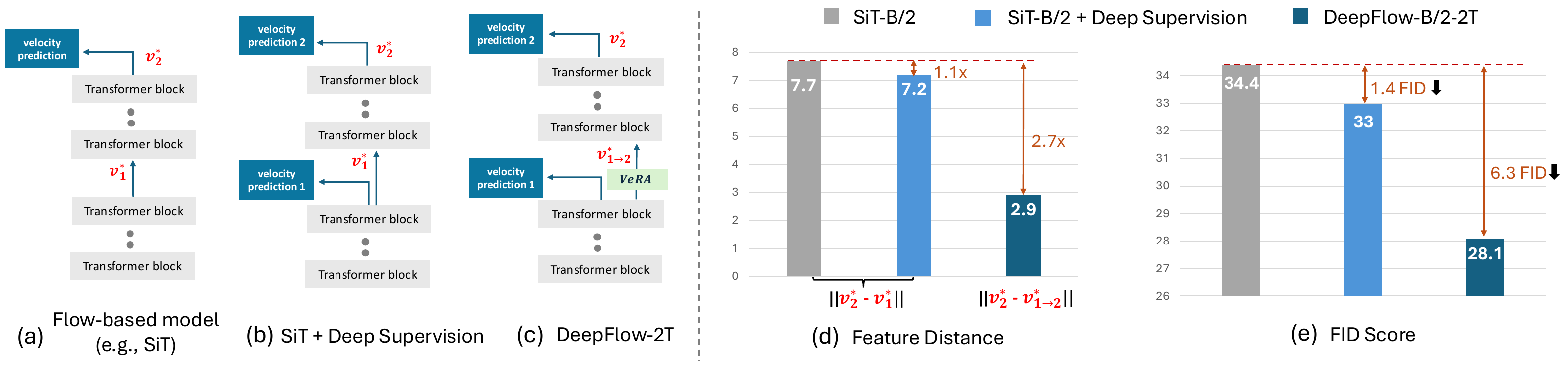}
\caption{\textbf{Importance of Internal Feature Alignment for Flow-Based Models.}
Our \modelname enhances the baseline flow-based model (a) by explicitly aligning intermediate velocity features with final layer features. As shown in (b), simply applying deep supervision reduces the feature distance between intermediate velocity $v^{*}_1$ (from middle 6th layer) and final $v^{*}_2$ (from 12th layer), improving FID scores (light blue bars in (d, e)). To further minimize this distance, we introduce the \blockname block, which refines deeply-supervised intermediate features into $v^{*}_{1\rightarrow2}$, more closely aligned with $v^{*}_2$ (dark blue bar in (d)). This leads to even better image generation quality (dark blue bars in (e)).
}
\label{fig:intro_feat}
\end{center}
\vspace{-3mm}
\end{figure*}

%% file: sections/1.introduction.tex
\section{Introduction}
\label{intro}

In the era of generative AI, it is indisputable that the strategy \textit{``denoising from noise''} has significantly propelled the advancement of visual generation. The processes of introducing and removing noise have given rise to two prominent families of generative models: diffusion-based models~\cite{ho2020denoising, rombach2022high, song2020score, song2020improved} and flow-based models~\cite{lipman2022flow, albergo2023stochastic, liu2022flow, lipman2024flow}. Diffusion-based models utilize a curved trajectory of diffusion forward process and denoise it back using an noise prediction. In contrast, flow-based models simply adopt the linear interpolation between noise and target signals, learning to predict velocity of interpolated noisy image under the principles of normalizing flows~\cite{chen2018neural, rezende2015variational}. Owing to these straightforward yet effective noising and denoising mechanisms, flow-based models have achieved state-of-the-art performance across numerous visual generation benchmarks~\cite{ma2024sit, yu2024representation, liu2023instaflow, esser2024scaling, jin2024pyramidal, gao2024lumina}.

Despite recent advancements, current fundamental flow-based models~\cite{ma2024sit, esser2024scaling} have largely overlooked the potential for enhancing their internal velocity representations. As illustrated in~\cref{fig:intro_feat}(a), SiT~\cite{ma2024sit}, a representative flow-based model, relies on sequentially stacked multi-layered transformer~\cite{vaswani2017attention} blocks to learn the velocity exclusively from the final layer. This approach under-utilizes the significance of intermediate velocity representations, leading to challenges such as slow training convergence and low performance~\cite{pernias2024wrstchen, yu2024representation}. Recently, to address this limitation, REPA~\cite{yu2024representation} aligns the internal velocity representation with external features from pre-trained self-supervised models (\eg, DINO~\cite{caron2021emerging, oquab2023dinov2}), resulting in better generative models with fewer training time needed.
However, relying solely on external self-supervised models overlooks the opportunity to internally rectify feature misalignment within transformer layers and fails to fully leverage the properties of flow-based models.
A natural question thus arises: \textit{Can flow-based models be improved by internally aligning velocity representations across transformer layers instead of relying on external models?}

We begin with a straightforward approach by incorporating deep supervision~\cite{lee2015deeply} within the transformer layers of flow-based models to enhance alignment. As depicted in~\cref{fig:intro_feat}(b), flow-based models can employ deep supervision across multiple velocity layers by partitioning transformer blocks into equal-sized branches, each trained to predict the same ground-truth velocity. 
This approach can align intermediate and final velocity features, as demonstrated by the reduced feature distance~\footnote{We compute euclidean distance between $v^{*}_1$ (from middle 6th layer) and $v^{*}_2$ (from final 12th layer) across all 250 timesteps using SDE while generating 50k samples on the ImageNet-256×256 dataset. This metric quantifies the alignment between intermediate and final layer velocity features.} in~\cref{fig:intro_feat}(d); SiT-B/2: \textbf{7.7} \textit{vs.} SiT-B/2 with deep supervision: \textbf{7.2}. This alignment, in turn, positively affects image generation performance in~\cref{fig:intro_feat}(e); SiT-B/2: \textbf{34.4 FID} \textit{vs.} SiT-B/2 with deep supervision: \textbf{33.0 FID}.

However, deep supervision alone is insufficient for achieving optimal alignment between intermediate and final layers, as intermediate layers exhibit a limited capacity for velocity prediction compared to the final layer. Motivated by this, we propose a redesigned flow-based model that explicitly aligns internal velocity representations while effectively integrating deep supervision—hereafter referred to as \textbf{\modelname}. It aims to refine deeply-supervised intermediate velocity features to be aligned for following branch as depicted in~\cref{fig:intro_feat}(c). To achieve this, \modelname consists of a lightweight block between adjacent branches, explicitly tailored to learn the mapping of velocity features from preceding branch to subsequent one.
This block, termed the \textbf{Ve}locity \textbf{R}efiner with \textbf{A}cceleration (\textbf{VeRA}) block, is specifically designed to model acceleration. It refines velocity features by conditioning on adjacent branches across different time steps. This process is guided by principles of second-order dynamics. To achieve this, we implement a simple MLP that takes previous velocity feature and is trained to generate an acceleration feature using the second-order ODE as visualized in \cref{fig:main_arc}.
Afterwards, we concatenate the previous velocity features with the computed acceleration features and apply a time-gap–conditioned adaptive layer normalization, ensuring aligned velocity features for the subsequent branch. Finally, we further refine these features by incorporating spatial information through cross-space attention, facilitating interaction between the refined velocity and spatial feature spaces.
We can observe that the velocity feature refined from \blockname block is significantly closer to the final output velocity feature in~\cref{fig:intro_feat}(d); SiT-B/2 with deep supervision: \textbf{7.2} \textit{vs.} DeepFlow-B/2-2T: \textbf{2.9}. It successfully leads to enhanced image quality as shown in~\cref{fig:intro_feat}(e); SiT-B/2 with deep supervision: \textbf{33.0 FID} \textit{vs.} DeepFlow-B/2-2T: \textbf{28.1 FID}.

Driven by feature alignment strategy for enhanced deep supervision, \modelname significantly improves both training efficiency and final image generation quality, all without dependence on external models. \cref{fig:eff_fid}(a) shows that \modelname-L/2-3T model with smaller number of parameters outperforms the SiT-XL/2 model after 80 epochs of training on ImageNet-256 benchmark~\cite{5206848}. Moreover, our \modelname-XL/2-3T model delivers performance comparable to the SiT-XL/2 model while reducing training time by eightfold, as shown in~\cref{fig:eff_fid}(b). It further improves generation quality by outperforming SiT-XL/2 only using half of training time.
For optimal image generation, we can seamlessly integrate feature alignment using external self-supervised model (\eg, DINO${v{2}}$~\cite{oquab2023dinov2}) and classifier-free guidance~\cite{ho2022classifier}, yielding better results on both ImageNet-256  and ImageNet-512 while requiring fewer training epochs needed.
Additionally, we performed extensive comparisons with a conventional flow-based model on the text-to-image generation benchmark using the MS-COCO dataset~\cite{lin2014microsoft} and GenEval benchmark~\cite{ghosh2024geneval}.



%% file: sections/2.related_works.tex
\section{Related Works}
\label{related}
\noindent \textbf{Generative Models with Denoising Transformers.}
Recent studies have advanced the field of visual generation by leveraging transformer~\cite{vaswani2017attention} architecture as a denoising model~\cite{song2020score, ho2020denoising, nichol2021improved, rombach2022high,liu2024alleviating,yang20241,ren2024flowar,ren2025beyond,he2025flowtok,ren2025grouping}. Specifically, U-ViT~\cite{bao2022all} and DiffiT~\cite{hatamizadeh2024diffit} integrate skip connections~\cite{ronneberger2015u} into  transformer-based backbones, whereas DiT~\cite{peebles2023scalable} demonstrates that a simple transformer-based diffusion network without skip connections can serve as a scalable and effective backbone for diffusion models.
Based on this simple architecture from DiT, SiT~\cite{ma2024sit} employs the principle of flow matching~\cite{lipman2022flow} and normalizing flows~\cite{rezende2015variational}, resulting in better image generation quality. Although both studies recognize that scaling laws hold as the number of transformer blocks increases, they overlook the role of internal feature representations across transformer layers.

\noindent \textbf{Feature Enhancement in Generative Models.}
Several approaches have sought to enhance internal representations in denoising transformers. For example, REPA~\cite{yu2024representation} improves generative modeling by aligning internal features in both diffusion and flow-based models with external representations from pre-trained self-supervised models like MAE~\cite{he2022masked} and DINO~\cite{caron2021emerging, oquab2023dinov2}. Similarly, VA-VAE~\cite{yao2025reconstruction} refines tokenizer~\cite{kingma2013auto} representations by incorporating external foundational models. However, relying solely on external models may overlook the self-correcting potential in addressing feature misalignment across intermediate layers.  Deep supervision~\cite{lee2015deeply} addresses this in classification tasks by providing multi-layer supervision that refines internal features for better discriminative performance. Inspired by this, we extend deep supervision to flow-based generative models, where the discriminative quality of velocity representations is critical.


\input{figures/main_figure}


%% file: figures/main_figure.tex
\begin{figure*}[!t]
\begin{center}
\includegraphics[width=0.8\linewidth]{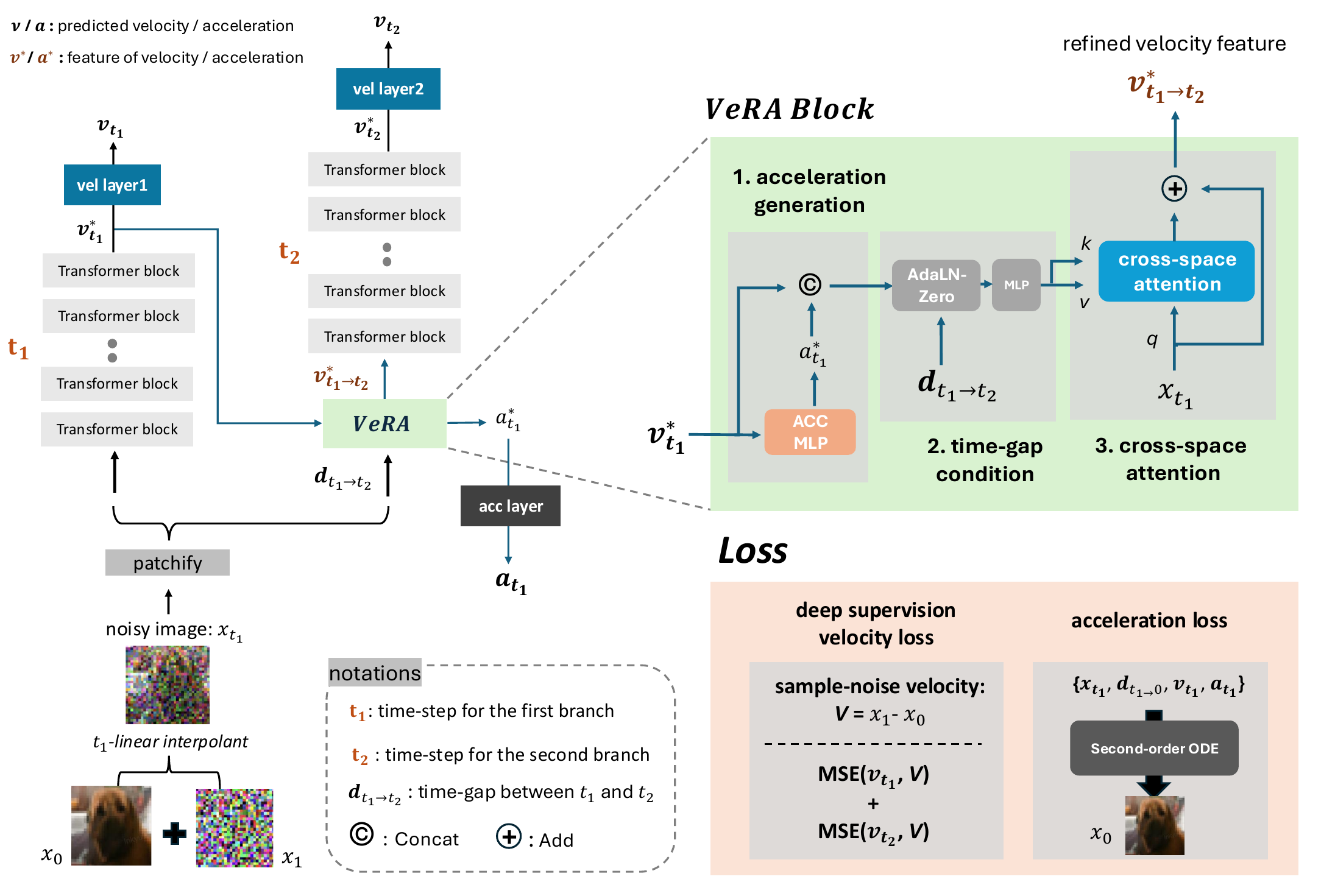}
\caption{\textbf{\modelname Architecture.} We introduce advanced deep supervision by partitioning transformer blocks into equal-sized branches and employing multiple velocity layers (dark blue boxes), enabling each branch to predict velocity at a distinct time-step. Then, \textcolor{green}{\blockname Block} is inserted between adjacent branches for explicit feature refinement. It consists of three sub-blocks:
1. acceleration generation: we design a simple MLP (\textcolor{orange}{ACC MLP}) to generate acceleration feature. It is trained with acceleration loss using second-order ODE function (\cref{eqn:acc}). Meanwhile, we concatenate previous velocity feature and computed acceleration feature for following sub-block. 2. time-gap condition: we modulate concatenated features using \textcolor{gray}{\textbf{AdaLN-Zero}} layer conditioned by time-gap. 3. \textcolor{bblue}{cross-space attention}: we design a novel cross-attention that integrates two features from different spaces, modulated velocity features from temporal dynamics and spatial features from original patchified image.
By leveraging time-gap conditioning between branches with \blockname block, \modelname enhances feature alignment and ultimately improves image generation quality.}
\label{fig:main_arc}
\end{center}
\vspace{-2mm}
\end{figure*}

%% file: sections/3.method.tex

\section{Method}
\label{method}
In this section, we first introduce the preliminaries on flow matching in~\cref{sec:preliminaries}, followed by a detailed presentation of the proposed method, \modelname, in~\cref{sec:deepflow}.

\subsection{Preliminaries}
\label{sec:preliminaries}

\noindent\textbf{Flow Matching.} Normalizing flows~\cite{chen2018neural} conceptualized time-dependent velocity field, $v : [0,1] \times \mathbb{R}^d \to \mathbb{R}^d$, which can provide flow map, $\phi : [0,1] \times \mathbb{R}^d$. This flow map aims to push-forward simple pure noise $z$ to target distribution, $x_{0}$. Flow matching~\cite{lipman2022flow} applied this finding to generative model by designing a neural network, $v_\theta(\mathbf{x}_t)$ that predicts the velocity of $\mathbf{x}_t$ at time-step $t$ with parameter $\theta$. Thus, it can generate target samples from pure gaussian noise using progressive denoising step with predicted velocity. To this end, during training, forward noising step is conducted with simple linear interpolation between prior noise ($\mathbf{x}_1 \sim \mathcal{N}(0,1)$) and target distribution ($\mathbf{x}_{0}$) as below:
\begin{equation}
\mathbf{x}_t = t \cdot \mathbf{x}_1 + (1 - t) \cdot \mathbf{x}_0,
\end{equation}
where $t \in [0, 1]$ denotes time-step used for interpolation coefficient.
Then, the flow-based method is learned to transform $v_\theta(\mathbf{x}_t)$ to be similar to corresponding ground-truth velocity $V=\mathbf{x}_1-\mathbf{x}_0$ with following objective function: 
\begin{equation}
\mathcal{L}(\theta) = \mathbb{E} \| v_\theta(\mathbf{x}_t) - V \|^2
\label{eqn:flow_eqn}
\end{equation}
Consistent with a conventional flow-based model~\cite{ma2024sit}, we use DiT Transformer~\cite{peebles2023scalable} as $v_\theta(\cdot)$. It comprises multiple transformer blocks that apply self-attention to the input tokens, followed by AdaLN-Zero modulation conditioned on the time step and class (or text), and a final velocity layer.
\cref{eqn:flow_eqn} is thus expressed as below:
\begin{equation}    
\mathcal{L}(\theta) = \mathbb{E} \| v_{\theta}(\mathbf{x}_t, t, c) - V \|^2,
\label{eqn:dit_eqn}
\end{equation}
where $t$ and $c$ are input time-step and class features.
Similarly, SiT~\cite{ma2024sit} adopts DiT Transformer while leveraging flow matching-based noising and denoising process. Depending on the number of transformer blocks and channel dimension used, SiT has four variants, 
$\text{SiT-\{S,B,L,XL\}}$.

\subsection{{{DeepFlow}}}
\label{sec:deepflow}

\noindent Our \modelname aims to exploit the potential of internal features across transformer blocks by enhancing feature alignment. It can be achieved by incorporating Deep Supervision (\cref{sec:deep_s}) into flow-based models and designing \blockname Block (\cref{sec:mit}) for explicit alignment between internal features. 

\subsubsection{Deep Supervision}
\label{sec:deep_s}

\modelname employs deep supervision~\cite{lee2015deeply} by inserting auxiliary velocity layers after selected intermediate transformer blocks. The corresponding deep supervision loss at these key transformer layers is defined as follows:
\begin{equation}
\begin{split}
    \mathcal{L}_{\text{deep}}(\theta) = \mathbb{E} \left[ \sum_{i=1}^{k} \beta^{i}(\| v_\theta^i(\mathbf{x}^{i}_t, t, c) - V \|^2) \right] \\
\end{split}
\label{eqn:deep}
\end{equation}
Here,  $k$ denotes the number of key layers, 
$v^{i}$ indicates the velocity prediction from the $i^{th}$ velocity layer following the corresponding transformer branch, and $\mathbf{x}^{i}_t$ denotes input features for the $i^{th}$ branch (defined as $\mathbf{x}_t$ for $i$=1, and velocity features from the previous branch otherwise).
$\beta^{i}$ represents the deep supervision coefficient. 
This loss encourages each velocity layer to produce outputs that closely match the target $V$.
This design enables our \modelname to support various configurations. For example, \modelname-\{$k$\}T  denotes a variant in which the transformer blocks are divided into
$k$ equal-sized Transformer branches, with each branch concluding with a velocity layer to facilitate deep supervision. For simplicity, we set $k$ as 2 for following explanation about \blockname block. 

\subsubsection{\blockname Block}
\label{sec:mit}
To enhance cross-layer deep supervision and improve feature alignment across different branches, we introduce a novel lightweight module called Velocity Refiner with Acceleration (\blockname) block.
This module explicitly aligns the deeply supervised velocity features between consecutive branches. 
We provide a step-by-step explanation of its overall architecture, as illustrated in ~\cref{fig:main_arc}.

\noindent \textbf{Branch Conditioned with Different Time-step.} During training, we deliberately differentiate time-steps conditioned on adjacent branches. Two branches correspond to two different time-steps as illustrated in ~\cref{fig:main_arc} with \textbf{\textcolor{lred}{$\text{t}_1$}} and \textbf{\textcolor{lred}{$\text{t}_2$}}, which effectively enables inserted \blockname block to be trained with second-order dynamics using time-gap. We first transform \cref{eqn:deep} for deep supervision as below to train each branch with its corresponding time-step.
\begin{equation}
    \begin{split}
\mathcal{L}_{\text{deep*}} = \mathbb{E} \left[  \sum_{i=1}^{2} \beta^{i}( \| v_\theta^i(\mathbf{x}^{i}_{{t}}, \textcolor{lred}{t_{i}}, c) - V \|^2) \right], \\
\mathbf{v}_{t_{1}} = v^{1}_{\theta}(\mathbf{x}^{1}_{t}, \textcolor{lred}{t_{1}}, c), \hspace{2mm} \mathbf{v}_{t_{2}} = v^{2}_{\theta}(\mathbf{x}^{2}_{t}, \textcolor{lred}{t_{2}}, c),\\
    \end{split}
    \label{eqn:deepstar}
\end{equation}
where $\mathbf{x}^1_{t}$ and $\mathbf{x}^2_{t}$ indicate initial noisy image $\mathbf{x}_{t_{1}}$ and previous velocity feature $\mathbf{v}^{*}_{t_{1}}$, respectively. 
It highlights that different time-steps are used for conditioning different branches to generate distinctive velocities.

\noindent \textbf{The \blockname Block Architecture.} 
The \blockname block is strategically placed between consecutive branches, refining deeply-supervised velocity features for use in the subsequent branch. The key operations include:

\vspace{0.5mm}
\noindent \underline{\textit{Acceleration Learning via Second-Order ODE Training:}} The primary goal of the \blockname block is to refine previous velocity features by incorporating acceleration information. To achieve this, we introduce a simple MLP block, termed ACC$\_$MLP, which projects the previous velocity feature ($\mathbf{v}_{t_{1}}^{*}$) to a higher dimension and then back to the original dimension, producing acceleration feature $\mathbf{a}_{t_{1}}^{*}$. 
\begin{equation}
    \mathbf{a}_{t_{1}}^{*} = \text{ACC}\_\text{MLP}(\mathbf{v}_{t_{1}}^{*})
\label{eqn:acc_mlp}
\end{equation}
Then, we can endow $\mathbf{a}_{t_{1}}^{*}$ with acceleration property using a second-order ordinary differential equation ($2^{nd}\text{-}ODE$) as following equation:
\begin{equation}
\begin{split}
    L_{acc} = \mathbb{E} \| 2^{nd}\text{-}ODE(\mathbf{x}_{t_{1}}, \mathbf{v}_{t_{1}}, \mathbf{a}_{t_{1}}, d_{t_{1}\rightarrow0}) - \mathbf{x}_0 \|^2, \\
    2^{nd}\text{-}ODE= \mathbf{x}_{t_{1}}\text{+}\mathbf{v}_{t_{1}}\odot d_{t_{1}\rightarrow0}\text{+}\frac{1}{2}\mathbf{a}_{t_{1}}\odot(d_{t_{1}\rightarrow0})^{2},
\end{split}
\label{eqn:acc}
\end{equation}
where $\mathbf{v}_{t_{1}}$ and $\mathbf{a}_{t_{1}}$ are outputs of velocity and acceleration layers from $\mathbf{v}_{t_{1}}^{*}$ and $\mathbf{a}_{t_{1}}^{*}$, reducing their dimension to image space.
$d_{t_{1}\rightarrow0}$ is time gap between time-steps of $t_{1}$ and 0. $\odot$ indicates Hadamard product for element-wise matrix multiplication. In this setup, the acceleration is learned in such a way that it aligns closely with the clean image representation ($\mathbf{x}_0$).

\vspace{0.5mm}
\noindent \underline{\textit{Feature Concatenation and Time-gap Conditioning:}} After computing the acceleration features ($\mathbf{a}_{t_{1}}^{*}$), we concatenate these with the original velocity features ($\mathbf{v}_{t_{1}}^{*}$). To enable this concatenated feature to be aware of time-gap, we apply a time-gap–conditioned adaptive layer normalization~\cite{peebles2023scalable} with a following MLP as below:
\begin{equation}
\begin{split}
    modulate(\mathbf{v}_{t_{1}}^{*}) = MLP(AdaLN\text{-}Zero(\\concat(\mathbf{v}_{t_{1}}^{*}, \mathbf{a}_{t_{1}}^{*}),T(d_{t_{1}\rightarrow t_{2}}))) 
\end{split}
\end{equation}
$d_{t_{1}\rightarrow t_{2}}$ denotes the time gap between $t_{1}$ and $t_{2}$ and passes through time embedder $T$.
This dynamically modulates the concatenated feature statistics based on the time difference, which steps forward refined velocity feature.

\vspace{0.5mm}
\noindent \underline{\textit{Spatial Information Integration via Cross-Attention:}} Beyond feature alignment with temporal property using different time-steps, the \blockname block also integrates spatial context by employing a cross-attention (\textit{CA}) mechanism. This mechanism facilitates interaction between two spaces: modulated velocity feature space from previous step and spatial feature space from an original patchified image as noted in following equation.
\begin{equation}
    \textcolor{tred}{\mathbf{v}_{t_{1}\rightarrow t_{2}}^{*}} = CA(modulate(\mathbf{v}_{t_{1}}^{*}), \mathbf{x}_{t_{1}}),
\end{equation}
where $modulate(\mathbf{v}_{t_{1}}^{*})$ is used for key and value, while $\mathbf{x}_{t_{1}}$ is used for query.
This approach using cross-space attention effectively highlights pertinent spatial features that may have been underrepresented in the pure temporal transformation, leading to final refined velocity feature.

\noindent In summary, \modelname incorporates deep supervision and \blockname block, 
which together enable training to align the internal velocity representation with 
\cref{eqn:deepstar} and \cref{eqn:acc}, as described below:
\begin{equation}
    L_{total} = L_{deep^{*}} + \lambda L_{acc},
    \label{eqn:balance}
\end{equation}
where $\lambda$ is a hyperparameter that balances the deep-supervised velocity loss and the acceleration loss.

%% file: sections/4.experiments.tex
\section{Experiments}
\label{exp}
In this section, we demonstrate the effectiveness of \modelname through extensive experiments. \cref{sec:implementation} details the implementation of \modelname. \cref{sec:main_res} presents the main results of \modelname on class-conditional image generation and text-to-image generation. Finally, comprehensive ablation studies are provided in \cref{sec:ablations}.

\subsection{Implementation Details}
\label{sec:implementation}

\input{tables/model_variants}

\noindent Our framework is implemented in PyTorch~\cite{paszke2019pytorch} and closely follows the flow matching and transformer setup introduced in SiT~\cite{ma2024sit}.
Specifically, we extract patchified features from raw images using the VAE~\cite{kingma2013auto} encoder pretrained on Stable Diffusion~\cite{rombach2022high}. For the transformer blocks, we divide them into equal-sized branches to enable deep supervision, with \blockname blocks inserted at key layers (see \cref{tab:mvar} for model configuration of \modelname). We explain more about used hyperparameters and implementation details in Appendix A.3.


\input{tables/model_compare_base_v2}

\noindent{\textbf{\modelname Training \& Evaluation.}}
 For training \blockname block in \modelname-\{$k$\}T, we employ $k$ distinct time-steps while constraining the maximum gap between consecutive branches to $\alpha$. 
 We empirically observed that assigning a low $\beta$ value (\eg, 0.2) to intermediate velocity predictions while maintaining a $\beta$ of 1.0 for final layer during training leads to improved generation quality.
 During evaluation, we condition all transformer layers on a single time-step, which still enables the refinement of velocity features across adjacent branches. We adopt basic training recipes and sampling strategies from SiT~\cite{ma2024sit} and REPA~\cite{yu2024representation} for fair comparisons.

\subsection{Main Results}
\label{sec:main_res}

\subsubsection{Class-conditional Image Generation}
ImageNet-1k~\cite{5206848} is widely used for class-conditional image generation benchmark. We provide the detail of its usage for training and sampling in Appendix A.7.

\input{tables/model_compare_xl_v2_24}
\input{figures/quality_epoch_main}

\noindent\textbf{Comparison with Flow-based Models.}
Using ImageNet-256×256, we compare our \modelname with the representative flow-based model SiT~\cite{ma2024sit} in \cref{tab:model_comparison_base} and \cref{tab:model_comparison_xl}, evaluated under the Base and XLarge configurations, respectively. In \cref{tab:model_comparison_base}, we consider two key comparison criteria: (i) the use of external self-supervised alignment (\textit{SSL align}), which leverages DINO$_{v1}$ or DINO$_{v2}$ as introduced in REPA~\cite{yu2024representation}, versus no SSL alignment;  (ii) the strategy for sampling time-steps during training, comparing \textit{uniform} sampling with \textit{lognormal} sampling, as also evaluated in SD3~\cite{esser2024scaling}. Without SSL align, \modelname-B/2-2T consistently outperforms SiT under both uniform and lognormal sampling strategies with a healthy margin, lowering 6.3 FID and 6.6 FID, respectively. When paired with SSL alignment using either DINO$_{v1}$ or DINO$_{v2}$, \modelname-B/2-2T further reduces the FID by an average of 3.0 points compared to SiT-B/2, achieving a remarkable generation quality with FID as low as 17.2. More surprisingly, \modelname-B/2-2T without SSL align can achieve comparable or better performance than SiT with SSL align from DINO$_{v1}$\footnote{DINO$_{v1}$ is used for this comparison since it was pretrained with ImageNet-1k, while DINO$_{v2}$ was pretrained on the additional dataset, LVD-142M~\cite{oquab2023dinov2}.} — for example,  28.3 FID \textit{vs.} 28.1 FID in uniform sampling, 24.4 FID \textit{vs.} 23.3 FID in lognormal sampling), demonstrating the effectiveness of \modelname in obviating the need for external feature alignment.
Experiments using the XLarge configuration (see \cref{tab:model_comparison_xl}) reveal two key observations. First, \modelname-XL/2-3T consistently outperforms SiT-XL/2—whether using SSL alignment or not—under lognormal sampling at the same training epochs (\eg, 80 and 200 epochs). Second, \modelname-XL/2-3T converges significantly faster than SiT-XL/2. For example, training \modelname-XL/2-3T for just 100 epochs without SSL alignment yields a 9.8 FID, matching the performance of SiT-XL/2 trained for 800 epochs. Furthermore, with 400 epochs, \modelname-XL/2-3T surpasses SiT-XL/2’s 800-epoch results. As illustrated in \cref{fig:qual_epochs_main}, our model also produces higher-quality images with fewer training epochs.





\input{tables/sota_compare_256_v2_24}
\noindent\textbf{Comparison with state-of-the-art Models.}
In addition to flow-based comparisons, we compare against state-of-the-art image generators—including autoregressive~\cite{sun2024autoregressive, luo2024open,yu2024randomized}, masked generative~\cite{ yu2023language,yu2024an,weber2024maskbit,kim2025democratizing}, diffusion-based~\cite{dhariwal2021diffusion, hoogeboom2023simple, peebles2023scalable}, and flow-based models~\cite{ma2024sit, yu2024representation}—using classifier-free guidance~\cite{ho2022classifier}. As shown in \cref{tab:sota_256} and \cref{tab:sota_512}, \modelname-XL/2-3T not only significantly outperforms existing flow-based models but also achieves competitive or superior performance compared to other state-of-the-art generators, all while training efficiently. For instance, \modelname-XL/2-3T delivers superior results on ImageNet-256 in only 400 epochs, and on ImageNet-512, it outperforms the corresponding SiT model with fewer epochs.
 



\input{tables/sota_compare_512_v2_24}

\input{tables/t2i_v2}
\subsubsection{Text-to-Image Generation}
Following REPA~\cite{yu2024representation}, we modify the architecture of MMDiT~\cite{esser2024scaling} by incorporating a flow matching objective, seamlessly integrating \modelname into MMDiT’s architecture. Both our model and SiT (MMDiT + flow matching) are trained on the MS-COCO~\cite{lin2014microsoft,chen2015microsoft} training set with a hidden dimension of 768 and 24 transformer layers while sampling time-steps from uniform distribution. We extensively conducted evaluation on MS-COCO validation set and GenEval~\cite{ghosh2024geneval} benchmark. For MS-COCO evaluation, we observe that \modelname outperforms in all of the metrics (FID, FD$_{DINO_{v2}}$~\cite{stein2023exposing}, IS, CLIP score). Furthermore, we show that \modelname significantly increase overall score of GenEval compared to SiT (check Appendiex A.5. for category-level performance in GenEval benchmark).









\input{tables/ablations_v4}

\input{tables/param_analysis_v2_color}

\subsection{Ablation Studies}
\label{sec:ablations}

\subsubsection{\modelname Components}
\vspace{-1mm}
\noindent \cref{tab:ablation:modules} summarizes the incremental improvements of \modelname over the baseline SiT-B/2~\cite{ma2024sit} (FID 34.4) using uniform sampling. Deep supervision on each branch reduces FID to 33.0 by aligning intermediate velocity features with the ground-truth. Adding a time-gap mechanism further lowers FID to 31.1 by enhancing robustness to time-step variations. An inter-layer acceleration pipeline within the \blockname block decreases FID to 29.9, and incorporating a cross-space attention module refines spatial fusion to achieve a final FID of 28.1. Overall, these enhancements yield a total FID improvement of 6.3, demonstrating the complementary benefits of each component in our framework.
\vspace{-2mm}

\vspace{-1mm}
\subsubsection{Time-gap} 
\vspace{-1mm}
\cref{tab:ablation:timegap} presents the impact of varying the maximum time-gap $\alpha$ between adjacent branches during training for \modelname. When $\alpha$ is set to 0.01, the model achieves its best FID of 28.1 under the base configuration. Additionally, we observe that the model's performance remains relatively stable across different values of $\alpha$, indicating that \modelname is not highly sensitive to variations in the time-gap parameter. 
\subsubsection{Acceleration Design}
\vspace{-1mm}
In \cref{tab:ablation:acc}, we explore how varying the balancing coefficient~$\lambda$ used in~\cref{eqn:balance} and the architecture of the acceleration MLP (ACC$\_$MLP) affect performance of \modelname. 
Considering the optimal $\lambda$ that works both in uniform sampling and lognormal sampling (marked with an asterisk), we choose $\lambda$ as 1.0.
We can also find out that expanding the ACC$\_$MLP capacity with deeper layers (the number of channels for ACC$\_$MLP: \{2048, 4096, 2048, 768\}) can increase the capability of acceleration, yielding further performance gains.

\vspace{-1mm}
\subsubsection{Efficiency \textit{vs.} Performance}
\label{sec:efficiecny}
\vspace{-1mm}
\cref{tab:param} compares various \modelname configurations (\modelname-B/2, \modelname-L/2, \modelname-XL/2) with baseline SiT models in terms of transformer depth, parameter count, GFLOPS, and FID. All models are trained for 80 epochs with lognormal sampling and evaluated using SDE (250 steps) on the ImageNet-256×256 benchmark without a classifier-free guidance. Two key observations emerge:
(i) \textbf{Efficient Performance Scaling}: Even smaller \modelname variants match larger SiT models. For example, \modelname-L/2-3T (18 layers, 433M parameters, 72 GFLOPS) achieves an FID of 13.3, comparable to SiT-XL/2 (675M parameters, 114 GFLOPS, 13.7 FID).
(ii) \textbf{Superior Performance}: \modelname variants consistently outperform comparable SiT models. \modelname-XL/2-3T reduces FID by 3.4 points relative to SiT-XL/2, while maintaining a similar computational cost. Moreover, \modelname scales effectively, as \modelname-XL/2-4T further improves performance down to 9.7 FID.


%% file: tables/model_variants.tex
\begin{table}[H]
  \centering
\scalebox{0.83}{
    \begin{tabular}{c|c|c|c}
     model  & depth & hidden dim &  key layer indices \\
      \shline
     \modelname-B/2-2T & 12 & 768 & \{6, 12\} \\
      \modelname-L/2-3T & 18 & 1024 & \{6, 12, 18\}  \\
       \modelname-XL/2-3T & 24 & 1152 & \{8, 16, 24\} \\
    \end{tabular}}
    \vspace{1mm}
\caption{
We propose three variants of \modelname (B, L, and XL), each differing in depth (\ie, the number of blocks) and channel dimensions. \modelname-$\{k\}$T denotes a model with $k$ key layers, where deep supervision is applied. These key layers are evenly distributed across the Transformer blocks.
}
\label{tab:mvar}
\vspace{-1mm}
\end{table}

%% file: tables/model_compare_base_v2.tex
\begin{table}[t]
\centering
\scalebox{0.555}{
\begin{tabular}{
    c|    
    c    
    c    
    c    
    | c    
    c    
    c    
}
model & epoch & SSL align & sampling & FID$\downarrow$& sFID$\downarrow$ & IS$\uparrow$ \\
\shline
\multirow{6}{*}{SiT-B/2~\cite{ma2024sit}} & 80 & \ding{55}  & uniform   & 34.4 & 6.5 & 43.7 \\
& 80 & \ding{55}  & lognormal & 29.7 &  6.2   & 51.0     \\
\cline{2-7}
& 80 & DINOv1  & uniform   & 28.3 & 6.6 & 53.8 \\
& 80& DINOv1 & lognormal & 24.4 & 6.3 & 62.1 \\
\cline{2-7}
 & 80 & DINOv2  & uniform   & 23.0 & 6.4 & 59.9 \\
& 80 & DINOv2  & lognormal & 20.4 & 6.3 & 72.7 \\
\midrule
\multirow{6}{*}{\modelname-B/2-2T} & 80 & \ding{55}  & uniform   & 28.1 \textcolor{blue}{(-6.3)} & 5.8 \textcolor{blue}{(-0.7)} & 49.8 \textcolor{blue}{(+6.1)}\\
 & 80 & \ding{55}  & lognormal & 23.1 \textcolor{blue}{(-6.6)} & 5.6 \textcolor{blue}{(-0.6)} & 60.3 \textcolor{blue}{(+9.3)} \\
\cline{2-7}
 & 80 & DINOv1  & uniform & 25.6 \textcolor{blue}{(-2.7)} & 6.4 \textcolor{blue}{(-0.2)}  & 56.6 \textcolor{blue}{(+2.8)} \\
 & 80 & DINOv1  & lognormal & 21.7 \textcolor{blue}{(-2.7)} & 6.1 \textcolor{blue}{(-0.2)} & 66.0 \textcolor{blue}{(+3.9)} \\
 \cline{2-7}
 & 80 & DINOv2  & uniform & 20.0 \textcolor{blue}{(-3.0)} & 6.2 \textcolor{blue}{(-0.2)}  & 70.2 \textcolor{blue}{(+10.3)} \\
& 80 & DINOv2  & lognormal & 17.2 \textcolor{blue}{(-3.2)} & 6.0 \textcolor{blue}{(-0.3)} & 77.8 \textcolor{blue}{(+5.1)} \\
\end{tabular}}
\caption{Quantitative comparisons of flow-based generative models under the \textbf{Base} configuration for class-conditional image generation on ImageNet-256×256 without classfier-free guidance~\cite{ho2022classifier}. \modelname-B/2-2T consistently outperforms SiT-B/2 across various settings, whether using SSL alignment~\cite{yu2024representation} or not, and under both uniform and lognormal sampling strategies.
}
\label{tab:model_comparison_base}
\end{table}

%% file: tables/model_compare_xl_v2_24.tex
\begin{table}[t]
\centering
\scalebox{0.53}{
\begin{tabular}{
    c|    
    c    
    c    
    c    
    | c    
    c    
    c    
}
model & epoch & SSL align & sampling & FID$\downarrow$ & sFID$\downarrow$ & IS$\uparrow$ \\
\shline
\multirow{6}{*}{SiT-XL/2~\cite{ma2024sit}} & 80 & \ding{55}  & lognormal & 13.8 & 5.0     & 91.0      \\
& 200 & \ding{55}  & lognormal & 10.2 & 5.3 & 113.6 \\
& 800 & \ding{55}  & lognormal & 9.8 & 7.3 & 128.2  \\
\cline{2-7}
& 80 & DINOv2  & lognormal & 7.2  & 5.1  & 134.3  \\
 & 200 & DINOv2 & lognormal & 6.2  & 5.3  & 149.6  \\
 & 800 & DINOv2 & lognormal & 5.7 & 6.4  & 171.0  \\
\midrule
\multirow{6}{*}{\modelname-XL/2-3T} & 80 & \ding{55}  & lognormal   & 10.3 \textcolor{blue}{(-3.4)}    & 4.8 \textcolor{blue}{(-0.2)}    & 105.2 \textcolor{blue}{(+14.2)}   \\
& 200 & \ding{55}  & lognormal & 7.8 \textcolor{blue}{(-2.4)}      & 4.8 \textcolor{blue}{(-0.5)}   & 127.3 \textcolor{blue}{(+13.7)}   \\
& \textbf{400} & \ding{55}  & lognormal  & 7.2 & 5.1 & 138.5 \\
\cline{2-7}
& 80 & DINOv2  & lognormal & 6.5 \textcolor{blue}{(-0.7)} & 4.9 \textcolor{blue}{(-0.2)}  & 134.9 \textcolor{blue}{(+0.6)}\\
& 200 & DINOv2  & lognormal & 5.4 \textcolor{blue}{(-0.8)} & 5.0 \textcolor{blue}{(-0.3)} & 151.6 \textcolor{blue}{(+2.0)} \\
& \textbf{400} & DINOv2  & lognormal & 5.0 & 5.2 & 162.0  \\
\end{tabular}}
\caption{Quantitative comparisons of flow-based generative models under the \textbf{XLarge} configuration for class-conditional image generation on ImageNet-256×256 without classfier-free guidance~\cite{ho2022classifier}. Our \modelname-XL/2-3T not only consistently delivers superior image generation quality compared to SiT-XL/2 when trained for an equivalent number of epochs, but it also converges significantly faster, requiring only half the training epochs (800 epochs $\rightarrow$ 400 epochs).
}
\label{tab:model_comparison_xl}
\end{table}

%% file: figures/quality_epoch_main.tex
\begin{figure}[t!]
\begin{center}
\includegraphics[width=0.95\linewidth]{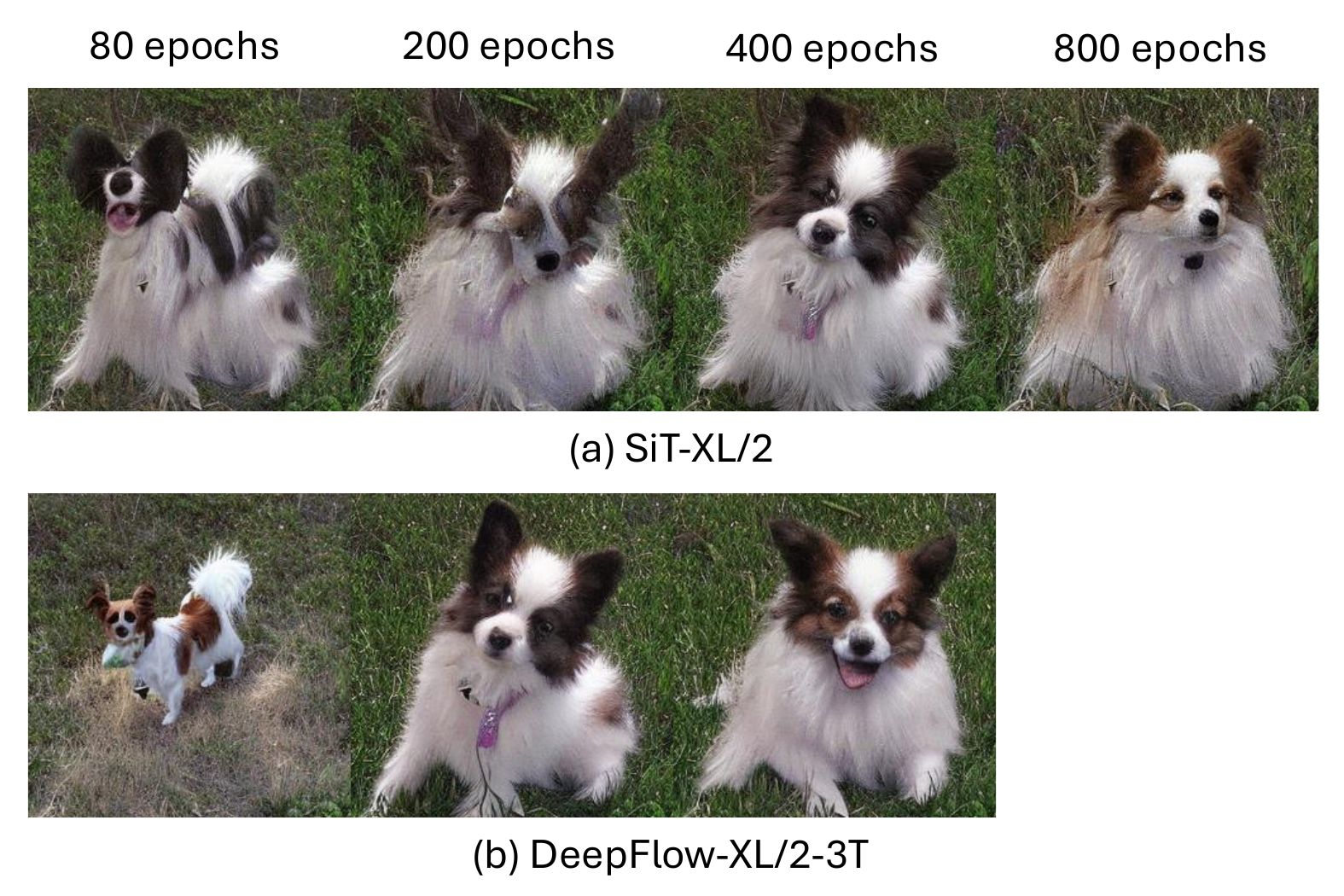}
\caption{
\textbf{Qualitative Comparison in Different Epochs.}
Images are generated from models trained in different epochs.
\modelname-XL/2-3T converges faster than SiT-XL/2 and produces high-quality samples even with fewer training epochs.
}
\label{fig:qual_epochs_main}
\end{center}
\vspace{-1mm}
\end{figure}

%% file: tables/sota_compare_256_v2_24.tex
\begin{table}[t]
    \centering
        \centering
        \resizebox{0.48\textwidth}{!}{ 
        \begin{tabular}{lccccc}
           \textbf{ImageNet-256$\times$256} & epoch & params & FID$\downarrow$ & sFID$\downarrow$ & IS$\uparrow$ \\
            \shline
            \multicolumn{6}{c}{\textit{Autoregressive Models}} \\
            \midrule
            LlamaGen-3B~\cite{sun2024autoregressive} & 50 & 3.1B & 3.05 & - & 222.3 \\
            Open-MAGVIT2-XL~\cite{luo2024open} & 350 & 1.5B & 2.33 & - & 271.8 \\
            \midrule
            \multicolumn{6}{c}{\textit{Masked Generative Models}} \\
            \midrule
            TiTok-S-128~\cite{yu2024an} & 200 & 287M & 1.97 & - & 281.8 \\
            MAGVIT-v2~\cite{yu2023language} & 270 & 307M & 1.78 & - & 319.4 \\
            \midrule
            \multicolumn{6}{c}{\textit{Diffusion-based Models}} \\
            \midrule
             ADM-U~\cite{dhariwal2021diffusion} & 400 & 554M & 3.94 & 6.14 & 186.7 \\
             Simple diffusion~\cite{hoogeboom2023simple} & 800 & - & 2.77 & - & 211.8 \\
             DiT-XL/2~\cite{peebles2023scalable} & 1400 & 675M & 2.27 & 4.60 & 278.2   \\
            \midrule
            \multicolumn{6}{c}{\textit{Flow-based Models}} \\
            \midrule
             SiT-XL/2~\cite{ma2024sit} & 1400 & 675M & 2.06 & 4.50 & 270.3  \\
             \modelname-XL/2-3T & \textbf{400}  & 681M & 1.97 & 4.39   & 264.7 \\
             \modelname-XL/2-3T & \textbf{600}  & 681M & 1.89 & 4.40   & 263.3 \\ \hline
             SiT-XL/2 + SSL align~\cite{yu2024representation} & 800 & 675M & 1.80 & 4.50 & 284.0 \\
             \modelname-XL/2-3T + SSL align & \textbf{400} & 681M & 1.77 & 4.44 & 271.3 \\
        \end{tabular}
        }
    \vspace{1mm}
    \caption{Comparison with state-of-the-art models in class-conditional image generation on ImageNet 256×256.
    }
    \label{tab:sota_256}
\end{table}


%% file: tables/sota_compare_512_v2_24.tex
\begin{table}[t]
    \centering
        \centering
        \resizebox{0.48\textwidth}{!}{ 
        \begin{tabular}{lccccc}
            \textbf{ImageNet-512$\times$512} & epoch & params & FID$\downarrow$ & sFID$\downarrow$ & IS$\uparrow$ \\
            \shline
            \multicolumn{6}{c}{\textit{Autoregressive Models}} \\
            \midrule
            GIVT-Causal-L+A~\cite{tschannen2025givt} & 500 & 1.7B & 2.92 & - & -  \\
            \midrule
            \multicolumn{6}{c}{\textit{Masked Generative Models}} \\
            \midrule
            TiTok-B-128~\cite{yu2024an} & 200 & 177M & 2.13 & -  & -  \\
            MAGVIT-v2~\cite{yu2023language} & 270 & 307M & 1.91 & - & 324.3  \\
            \midrule
            \multicolumn{6}{c}{\textit{Diffusion-based Models}} \\
            \midrule
             LDM-4-G~\cite{rombach2022high} & 167 & 400M & 3.60 & - & 247.7  \\
             DiT-XL/2~\cite{peebles2023scalable} & 600 & 676M & 3.04 & - & -  \\
            \midrule
            \multicolumn{6}{c}{\textit{Flow-based Models}} \\
            \midrule
             SiT-XL/2~\cite{ma2024sit} & 600 & 676M & 2.62 & 4.18  & 252.2 \\
             \modelname-XL/2-3T & \textbf{200}  & 682M &  2.59 & 4.23 & 231.1  \\ 
             \modelname-XL/2-3T & \textbf{400}  & 682M & 2.22 & 4.27 & 249.4  \\ \hline
             SiT-XL/2 + SSL align~\cite{yu2024representation} & 200 & 676M & 2.08  & 4.19 & 274.6 \\
             \modelname-XL-3T + SSL align & \textbf{160}  & 682M & 2.03  & 4.29  & 248.8  \\
             \modelname-XL-3T + SSL align & \textbf{200}  & 682M & 1.96 & 4.28  & 260.3  \\
        \end{tabular}
        }
    \vspace{1mm}
    \caption{Comparison with state-of-the-art models in class-conditional image generation on ImageNet 512×512.}
    \label{tab:sota_512}
\end{table}

%% file: tables/t2i_v2.tex
\begin{table}[t]
\centering
\scalebox{0.66}{
\begin{tabular}{
    c|    
    c    
    | c    
    c    
    c    
    c
    | c
}
 \multirow{2}{*}{model} & \multirow{2}{*}{SSL align}  & \multicolumn{4}{c|}{\underline{\textbf{MS-COCO Eval}}} & \underline{\textbf{GenEval}} \\
 & & FID$\downarrow$ & FD$_{DINOv2}$$\downarrow$ & IS$\uparrow$ & CLIP$\uparrow$ & Overall$\uparrow$ \\
\shline
\multirow{2}{*}{SiT~\cite{ma2024sit}}  & \ding{55}  & 5.38  & 408.27  & 31.1 & 0.2954 & 0.2672      \\
& DINOv2 & 4.24  & 274.19  & 34.4 & 0.2979 & 0.3166   \\
\midrule
\multirow{2}{*}{\modelname-3T} & \ding{55} & 5.32  &  343.60 &  33.0  & 0.2974 & 0.2957      \\
& DINOv2 & 4.18 & 216.20  &  35.9 & 0.3010 & 0.3458     \\
\end{tabular}}
\vspace{1mm}
\caption{Quantitative comparison on text-to-image generation. To ensure a fair comparison, both SiT~\cite{ma2024sit} and \modelname utilize 24 layers of transformer blocks and use same MS-COCO~\cite{lin2014microsoft} training set. Then, they are both compared for MS-COCO evaluation and GenEval~\cite{ghosh2024geneval} benchmark (please refer to category-level performance of GenEval in Appendix). We reproduced MS-COCO evaluation performance of SiT variants.
}
\label{tab:t2i}
\end{table}

%% file: tables/ablations_v4.tex

\begin{table*}[t]\vspace{-3mm}
\subfloat[\textbf{Component Ablation}: study on effectiveness of components for \modelname. \label{tab:ablation:modules}]{
\vspace{4mm}
\tablestyle{1.5pt}{1.05}\begin{tabular}{l|x{22}}
 \scriptsize model & FID$\downarrow$  \\
\shline
 \scriptsize SiT-B/2~\cite{ma2024sit} & 34.4  \\\hline
 \scriptsize + deep supervision  & 33.0  \\
 \scriptsize + time-gap  & 31.1 \\
 \scriptsize + inter-layer acceleration  & 29.9 \\
 \scriptsize \baseline{+ cross-space attention}  & \baseline{28.1} 
\end{tabular}}\hspace{3mm}
\subfloat[\textbf{Time-gap}: ablation on maximum time-gap ($\alpha$) between adjacent branches used during training. \label{tab:ablation:timegap}]{
\vspace{5mm}
\tablestyle{1.5pt}{1.0}\begin{tabular}{c|x{22}x{22}}
 \scriptsize model &  
\scriptsize $\boldsymbol{\alpha}$ & FID$\downarrow$  \\
\shline
 \scriptsize \multirow{5}{*}{\modelname-B/2-2T}  & 0.1 & 28.9 \\
& 0.05  & 28.6  \\
 & \baseline{0.01} & \baseline{28.1}  \\
  & 0.005  & 28.3  \\
  & 0.001  & 28.5  
\end{tabular}}\hspace{3mm}
\subfloat[\textbf{Acceleration Design}: ablation on $\lambda$ and number of channels for ACC\_MLP \label{tab:ablation:acc}. * indicates the model trained with lognormal sampling of time-steps.]{
\vspace{-0.3mm}
\tablestyle{0.92pt}{0.83}\begin{tabular}{c|c|c|x{22}}
 \scriptsize model & \scriptsize\textbf{$\lambda$} & \scriptsize\textbf{ACC\_MLP} 
 & FID$\downarrow$ \\
\shline
 \multirow{6}{*}{\modelname-B/2-2T}  & 1.0 & \{2048, 768\} & 30.4  \\ \cline{2-4}
 & 0.5 & \multirow{5}{*}{\{\textbf{2048, 4096, 2048, 768}\}} & 27.8 \\
  & 0.75 & & 28.1 \\
  & \baseline{1.0} & & \baseline{28.1} \\
  & 1.25 &  & 28.5 \\
  & 1.5 &  & 28.8 \\\hline
  \multirow{2}{*}{\modelname-B/2-2T$^{*}$} & 0.5 & \multirow{2}{*}{\{\textbf{2048, 4096, 2048, 768}\}} & 23.5 \\
  & \baseline{1.0} & & \baseline{23.1}
\end{tabular}}\vspace{2mm}\\
\caption{\textbf{\modelname Ablations}.
}
\label{tab:ablations}\vspace{-3mm}
\end{table*}

%% file: tables/param_analysis_v2_color.tex
\begin{table}[t!]
\centering
\scalebox{0.68}{
    \begin{tabular}{c|c|c|c|x{22}}
     model  & depth &  params & GFLOPS & FID$\downarrow$ \\
      \shline
      SiT-B/2~\cite{ma2024sit}   & \lighterred{12}       & \lighterred{130M}  &  \lighterred{24} & \lighterred{29.7}  \\ \midrule
      SiT-L/2~\cite{ma2024sit}    & \lightred{24}       & \lightred{458M}  & \lightred{80} &  \lightred{16.1} \\ \midrule
      SiT-XL/2~\cite{ma2024sit}    & \darkred{28}       & \darkred{675M}  & \darkred{114} & \darkred{13.7} \\
      \midrule \midrule
      \multirow{2}{*}{\modelname-B/2-2T} & 10 & 144M & 24 & 27.3 \\
     & \lighterblue{\textbf{12}} & \lighterblue{\textbf{166M}} & \lighterblue{\textbf{28}} & \lighterblue{\textbf{23.1}} \\
      \midrule
      \multirow{3}{*}{\modelname-L/2-2T} & 20 & 431M & 72 & 13.9 \\
     & 22 & 469M & 78 & 13.3  \\
     & 24 & 507M & 84 &  12.8 \\
     \midrule
      \multirow{4}{*}{\modelname-XL/2-2T} & 22 & 588M & 97 & 11.9 \\
     & 24 & 636M & 106 & 11.7 \\
       & 26 & 683M & 114 & 11.1 \\
       & 28 & 731M & 122 & 11.1 \\
      \midrule \midrule
      \multirow{3}{*}{\modelname-L/2-3T} & \lightblue{\textbf{18}} &  \lightblue{\textbf{433M}} & \lightblue{\textbf{72}} & \lightblue{\textbf{13.3}}   \\
       & 21 &  490M & 82 & 12.0  \\
       & 24 & 547M & 92 &  11.9 \\
       \midrule
      \multirow{4}{*}{\modelname-XL/2-3T} & 18 &  538M & 89 & 12.4  \\
       & 21 &  609M & 101 & 10.9  \\
       & \darkblue{\textbf{24}} & \darkblue{\textbf{681M}} & \darkblue{\textbf{113}} & \darkblue{\textbf{10.3}} \\
      & 27 & 753M & 125 & 10.0 \\
      \midrule \midrule
      \modelname-XL/2-4T   & 28   &  822M & 137 & 9.7 \\
    \end{tabular}}
    \vspace{2mm}
\caption{\textbf{Tradeoff between Efficiency and Performance of  \modelname.} All of the experiments, including the reproduced, better baseline SiT variants, utilize lognormal sampling with 80 epochs of training in ImageNet-256×256 and are evaluated using SDE 250 steps without a classifier-free guidance~\cite{ho2022classifier}. Best viewed in color.}
\label{tab:param}
\vspace{-2mm}
\end{table}

%% file: sections/5.conclusion.tex
\section{Conclusion}
\label{con}
We introduced \modelname, a novel flow-based generative model that enhances internal velocity representations via deep supervision and explicit feature alignment using proposed \blockname block. Our extensive experiments show that \modelname not only dramatically improves training efficiency of flow-based models while achieving competitive performance on various image generation benchmarks.
We believe this work lays a strong foundation for efficient and high performing flow-based models.


%% file: sections/6.appendix.tex
\appendix
\section{Appendix}
In the appendix, we provide additional information as listed below:

\begin{itemize}
    \item \cref{sec:add_feat_dist} provides additional analysis on feature distance.
    \item \cref{sec:design} provides the design choices of \blockname block.
    \item \cref{sec:hyper} provides the details of hyperparameters and implementations.
    \item \cref{sec:sensitivity} provides the analysis on different number of samplings.
    \item \cref{sec:geneval_bench} provides the category-level results on GenEval benchmark.
    \item \cref{sec:quality} provides additional qualitative results on image generation benchmarks.
    \item \cref{sec:dataset} provides the details of datasets and metrics used for experiments.
    \item \cref{sec:dis_lim} provides the discussion and limitation on our method.
\end{itemize}

\subsection{Additional Analysis on Feature Distance}
\label{sec:add_feat_dist}
To further investigate the alignment of velocity features across layers, we analyze the feature distance between all intermediate layers and the final layer. This extends the analysis from Figure 2 in the main paper, where only the distance between the features of key layer and the final layer was measured. By examining the full layer-wise distance trends, we can better understand how intermediate representations evolve toward the final velocity feature. As shown in ~\cref{fig:feature_layer}, \modelname consistently reduces the feature distance across layers, ensuring a smooth progression toward the final layer representation. Even with deep supervision and the integration of the \blockname block at the key layer (6th), \modelname maintains effective feature alignment throughout the network.

\input{figures/layer_feature_dist}

\input{figures/mit_design}
\subsection{Design Choices of \blockname Block}
\label{sec:design}
We present the design choices for the \blockname block, a core component of our \modelname, as illustrated in~\cref{fig:mitdesign}. Both designs leverage acceleration to refine preceding velocity features using an ACC MLP, adaptive layer normalization, and cross-space attention. The design in the left panel is motivated by first-order dynamics using addition of velocity and modulated acceleration. Specifically, $a^{*}_{t_{1}}$ from ACC MLP is modulated by $d_{t_{1}\rightarrow t_{2}}$, and added with $v^{*}_{t_{1}}$. In base configuration, this approach achieves 29.3 FID, outperforming SiT-B/2~\cite{ma2024sit} (34.4 FID)—but underperforms compared to the proposed \blockname block (in the right panel of ~\cref{fig:mitdesign}). Although the left design adheres more closely to first-order dynamics, modulating acceleration alone with a time-gap is insufficient to fully adjust the preceding velocity features. In contrast, our proposed \blockname optimizes feature alignment by modulating a concatenation of velocity and acceleration features, which results in superior generation performance (28.1 FID).

\subsection{Hyperparameters and Implementations}
\label{sec:hyper}

We provide detailed explanation about hyperparameters and implementations used for \modelname in following orders.

\input{tables/t2i_geneval_v2}
\begin{itemize}
    \item \textbf{Image Encoder: }We utilize VAE~\cite{kingma2013auto} encoder to pre-compute the latent feature of input as what SiT~\cite{ma2024sit} and REPA~\cite{yu2024representation} did. The checkpoint of VAE encoder is from \textit{stability/sd-vae-ft-ema}, which was pre-trained in Stable Diffusion~\cite{rombach2022high}. Then, we flatten the latent features with patch size of 2.
    \item \textbf{Transformer Blokcs: }We employ same setting of DiT~\cite{peebles2023scalable} to construct transformer blocks including branches of pre- and post-\blockname block. What we differentiate is we condition them with different time-step during training to train \blockname block with time-gap prior. We set time-gap to be same or under 0.01 as we ablated in main paper.
    \item \textbf{\blockname Block: }As the first core block of \blockname block, ACC MLP consists of 4 linear layers with SiLU~\cite{elfwing2018sigmoid} activation. Then, adaptive layer normalization with zero-initialization for final linear inputs time-gap to produce scale and shift for concatenated features of velocity and acceleration. For final part, cross-space attention module is performed with layer pre-norm modulated velocity feature space (key and value) and pre-norm spatial feature space (query).
    \item \textbf{Optimizer and Training: }To optimizer baselines and our \modelname, we utilize AdamW~\cite{loshchilov2017decoupled} with constant learning rate of 1e-4, ($\beta_{1}$, $\beta_{2}$) = (0.9, 0.999) without weight decay and train the models with batch size of 256. 
    For faster training, all of the experiments including \modelname and baselines were conducted using Pytorch Accelerate~\cite{accelerate} pipeline with mixed-precision (fp16), and A100 GPUs. 
    \item \textbf{SSL Alignment: }As demonstrated in our main paper, we employ an SSL encoder for additional feature alignment, following the approach of REPA~\cite{yu2024representation}. Unlike REPA, which aligns a manually selected key layer with the SSL encoder, we incorporate external alignment after the output of each \blockname block in a more unified manner. For instance, in \modelname-B/2-2T with SSL alignment, the refined features produced by the \blockname block are further aligned using either DINO$_{v1}$ or DINO$_{v2}$. In \modelname-XL/2-3T with SSL alignment, DINO$_{v2}$ is applied twice—once after each \blockname block. Notably, we also experimented with applying SSL alignment twice in the original SiT~\cite{ma2024sit}, but this did not lead to any performance improvement.
    \item \textbf{Inference (sampling): } In line with SiT~\cite{ma2024sit} and REPA~\cite{yu2024representation}, we adopt an SDE sampling strategy and perform 250 steps to ensure a fair comparison. We also search for the optimal classifier-free guidance (CFG) scale during the evaluation of \modelname. As shown in \cref{tab:cfg_eval}, \modelname-XL/2-3T without SSL alignment achieves its best FID performance at a CFG scale of 1.325, whereas \modelname-XL/2-3T with SSL alignment reaches optimal performance at a CFG scale of 1.3.
\end{itemize}

\subsection{Sensitivity to Different Number of Samplings}
\label{sec:sensitivity}
\cref{fig:sampling_steps} illustrates the performance sensitivity of our \modelname model to varying numbers of sampling steps and highlights its robustness compared to SiT~\cite{ma2024sit}. Notably, \modelname maintains stable performance across sampling steps ranging from 50 to 250 (with a mean FID of 11.1 and a standard deviation of 1.21), suggesting that it is less sensitive to changes in the number of steps than SiT~\cite{ma2024sit}, which exhibits a higher mean FID of 14.8 and a standard deviation of 1.52. Furthermore, \modelname surpasses SiT’s performance at 250 steps even when using only 50 steps. These results underscore the efficiency of \modelname: it not only reduces computational cost by requiring fewer steps, but it also delivers superior overall performance. 

\input{tables/cfg_eval}

\input{figures/sample_steps}
\subsection{More Detailed Results on GenEval Benchmark}
\label{sec:geneval_bench}

~\cref{tab:t2i_geneval} presents a zero-shot text-to-image generation comparison between \modelname-2/3T and SiT~\cite{ma2024sit}, both using 24 transformer layers on GenEval benchmark~\cite{ghosh2024geneval}. Overall, \modelname outperforms SiT across most categories, including Single Object, Two Object, and Counting, indicating better handling of object complexity and quantity. \modelname also achieves higher scores in color-related tasks (Colors and Color Attributes) and positioning, demonstrating more accurate object placement and color fidelity. Moreover, incorporating SSL alignment (e.g., DINOv2) benefits both models but consistently maintains \modelname’s performance advantage.


\subsection{Additional Qualitative Results}
\label{sec:quality}
\begin{figure*}[t]
\begin{center}
\includegraphics[width=1.01\linewidth]{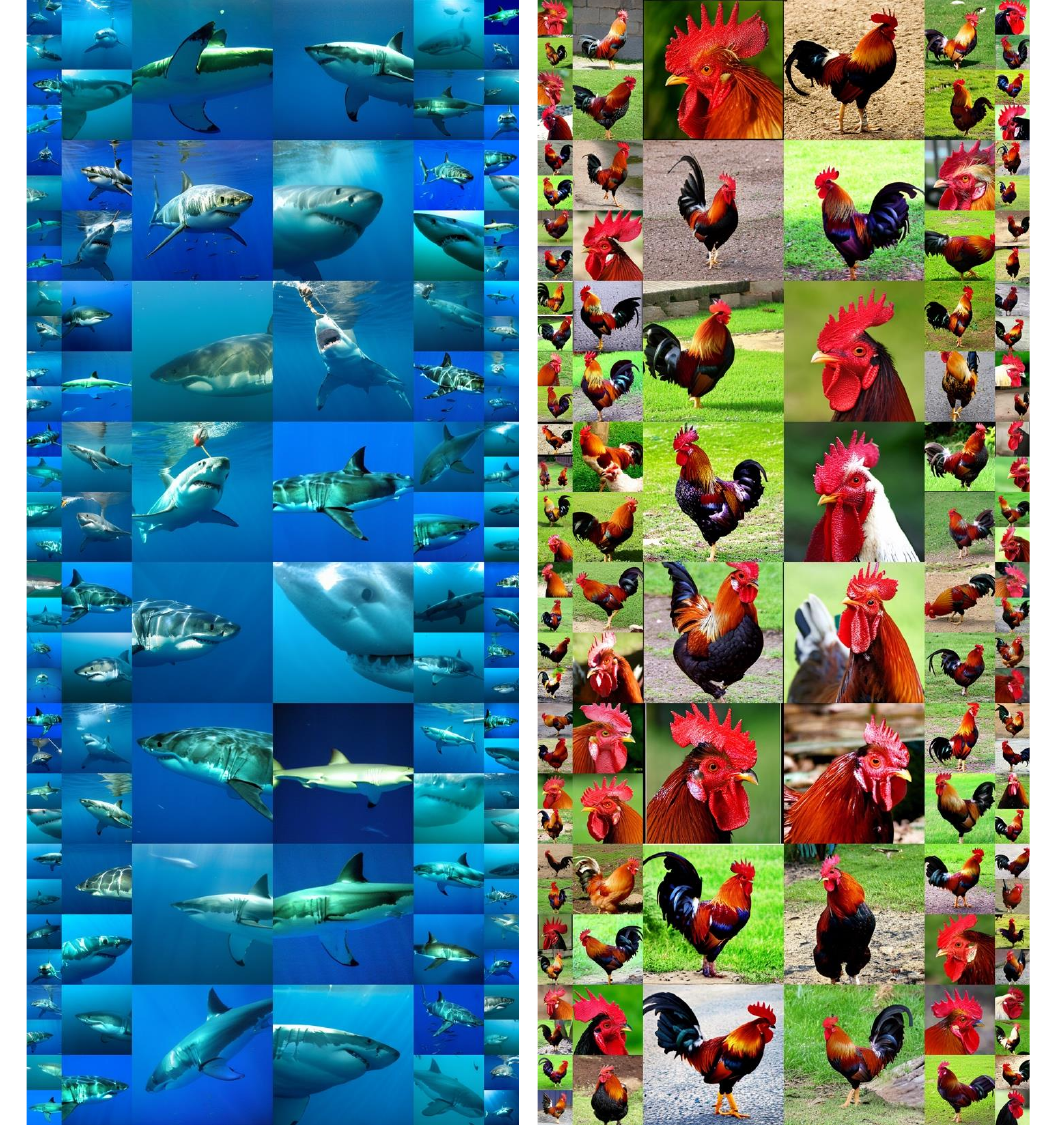}
\vspace{1mm}
\centering
\caption{
  \begin{tabular}{@{}l@{}}
\textbf{Uncurated} 256×256 \textbf{\modelname-XL/2-3T Samples (1).} Classifier-free guidance scale = 4.0. \\
(Left): Class = ``white shark" (2) \\
(Right): Class = ``cock" (7)
\end{tabular}
}
\label{fig:qc1}
\end{center}
\end{figure*}

\begin{figure*}[t]
\begin{center}
\includegraphics[width=1.01\linewidth]{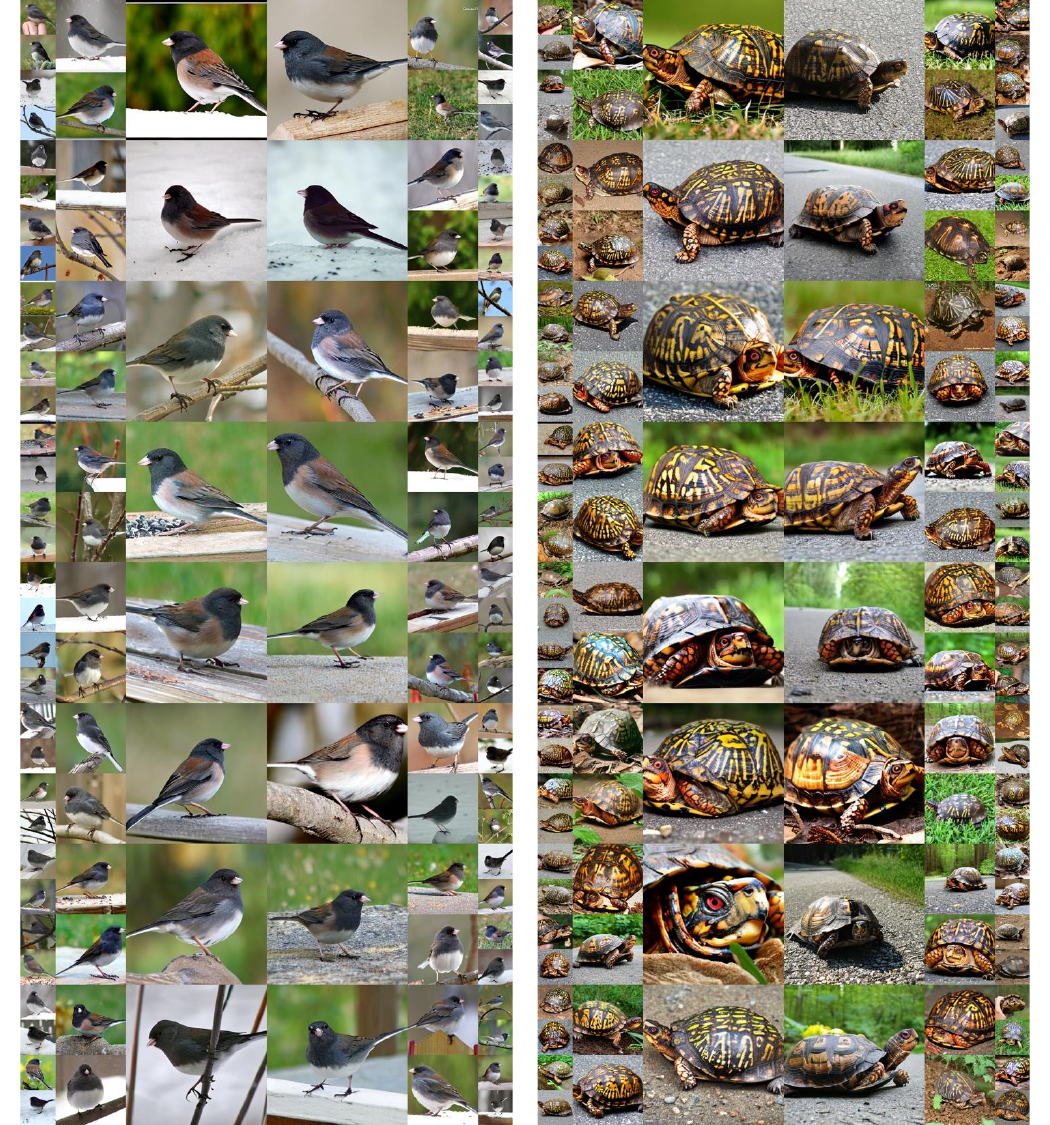}
\vspace{1mm}
\centering
\caption{
  \begin{tabular}{@{}l@{}}
\textbf{Uncurated} 256×256 \textbf{\modelname-XL/2-3T Samples (2).} Classifier-free guidance scale = 4.0. \\
(Left): Class = ``snowbird" (13) \\
(Right): Class = ``box turtle" (37)
\end{tabular}
}
\label{fig:qc2}
\end{center}
\end{figure*}

\begin{figure*}[t]
\begin{center}
\includegraphics[width=1.01\linewidth]{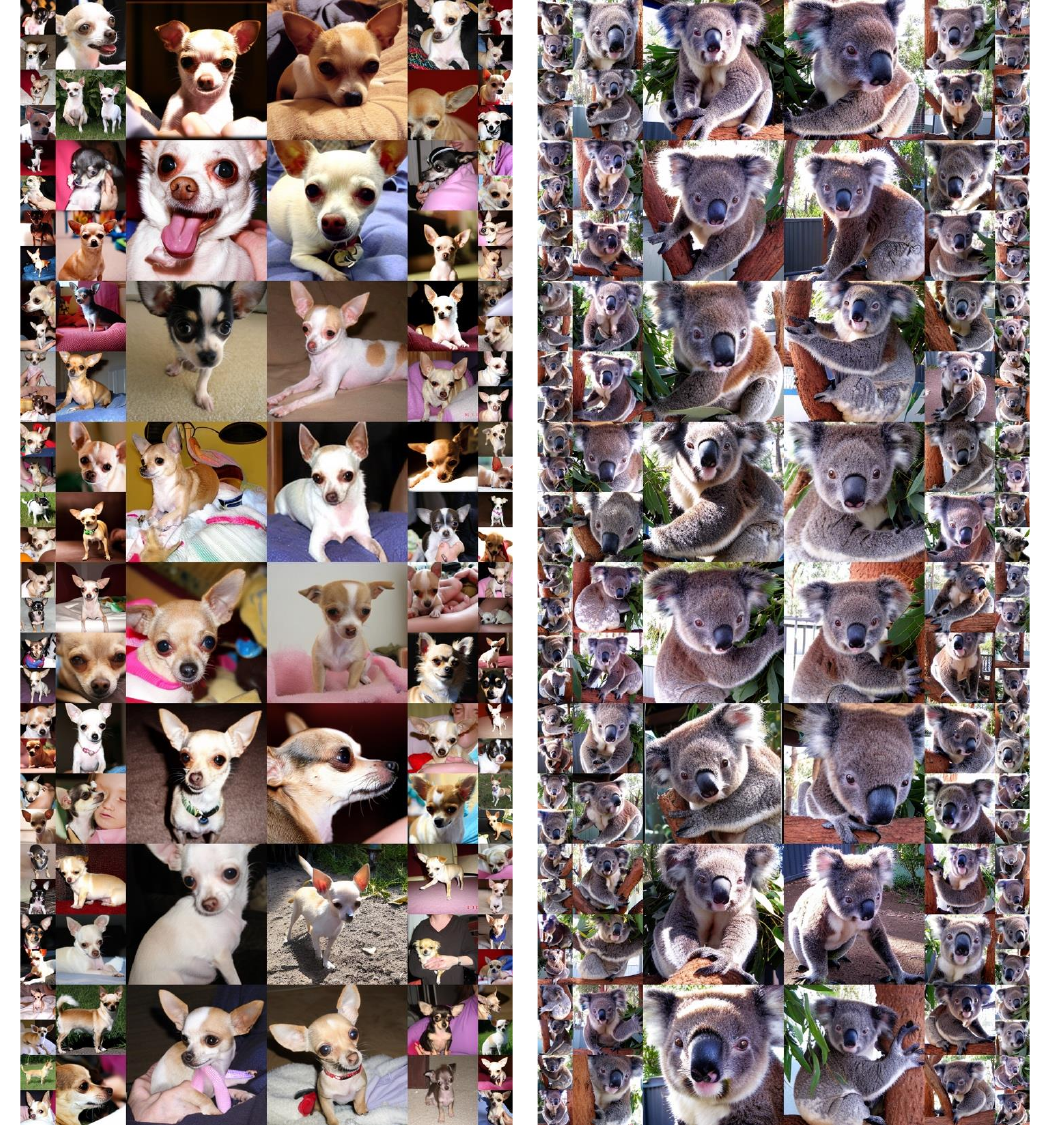}
\vspace{1mm}
\centering
\caption{
  \begin{tabular}{@{}l@{}}
\textbf{Uncurated} 256×256 \textbf{\modelname-XL/2-3T Samples (3).} Classifier-free guidance scale = 4.0. \\
(Left): Class = ``Chihuahua" (151) \\
(Right): Class = ``koala" (105)
\end{tabular}
}
\label{fig:qc3}
\end{center}
\end{figure*}

\begin{figure*}[t]
\begin{center}
\includegraphics[width=1.01\linewidth]{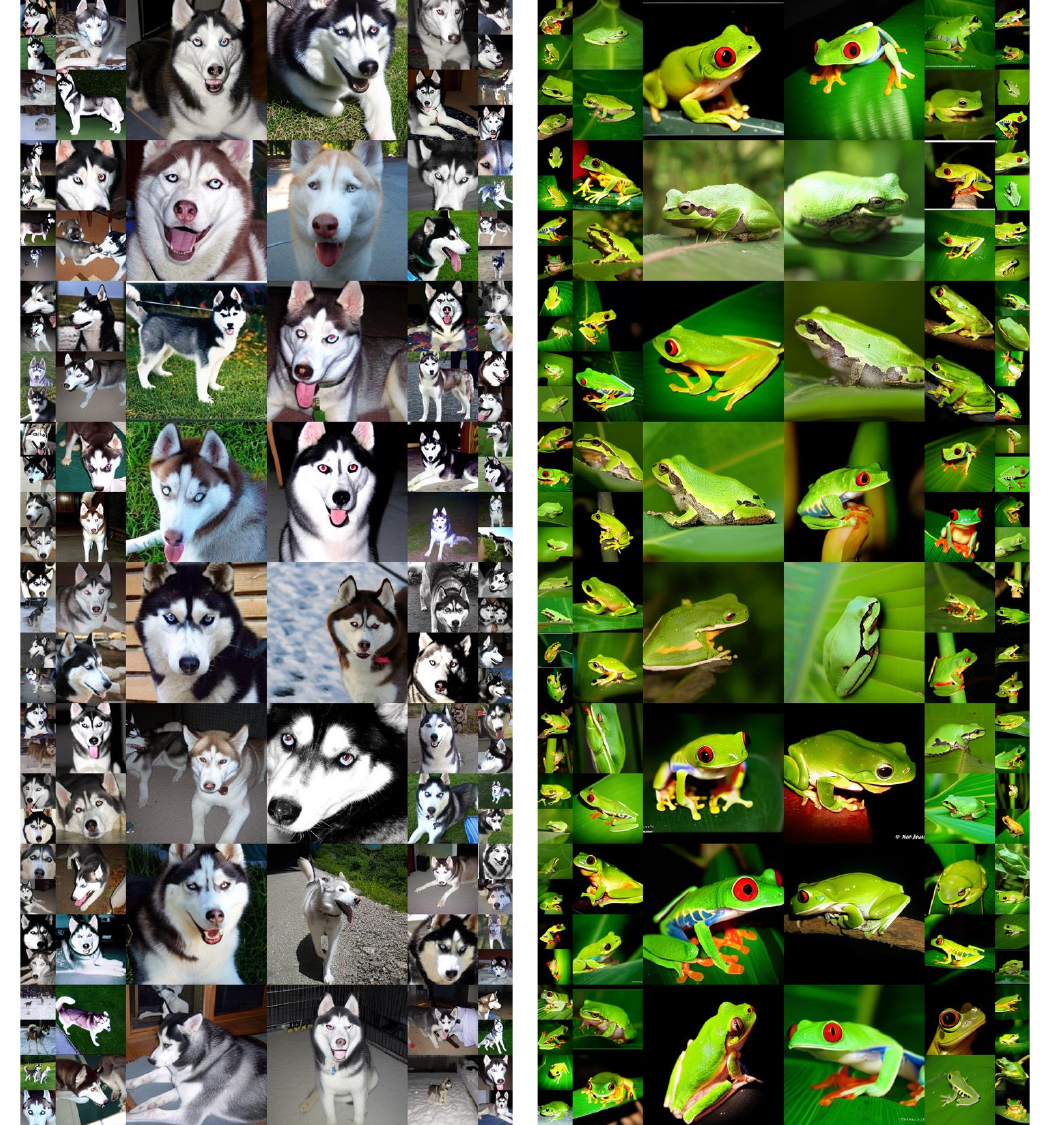}
\vspace{1mm}
\centering
\caption{
  \begin{tabular}{@{}l@{}}
\textbf{Uncurated} 256×256 \textbf{\modelname-XL/2-3T+SSL align~\cite{yu2024representation} Samples (1).} Classifier-free guidance scale = 4.0. \\
(Left): Class = ``Siberian husky" (250) \\
(Right): Class = ``tree frog" (31)
\end{tabular}
}
\label{fig:qc4}
\end{center}
\end{figure*}

\begin{figure*}[t]
\begin{center}
\includegraphics[width=1.01\linewidth]{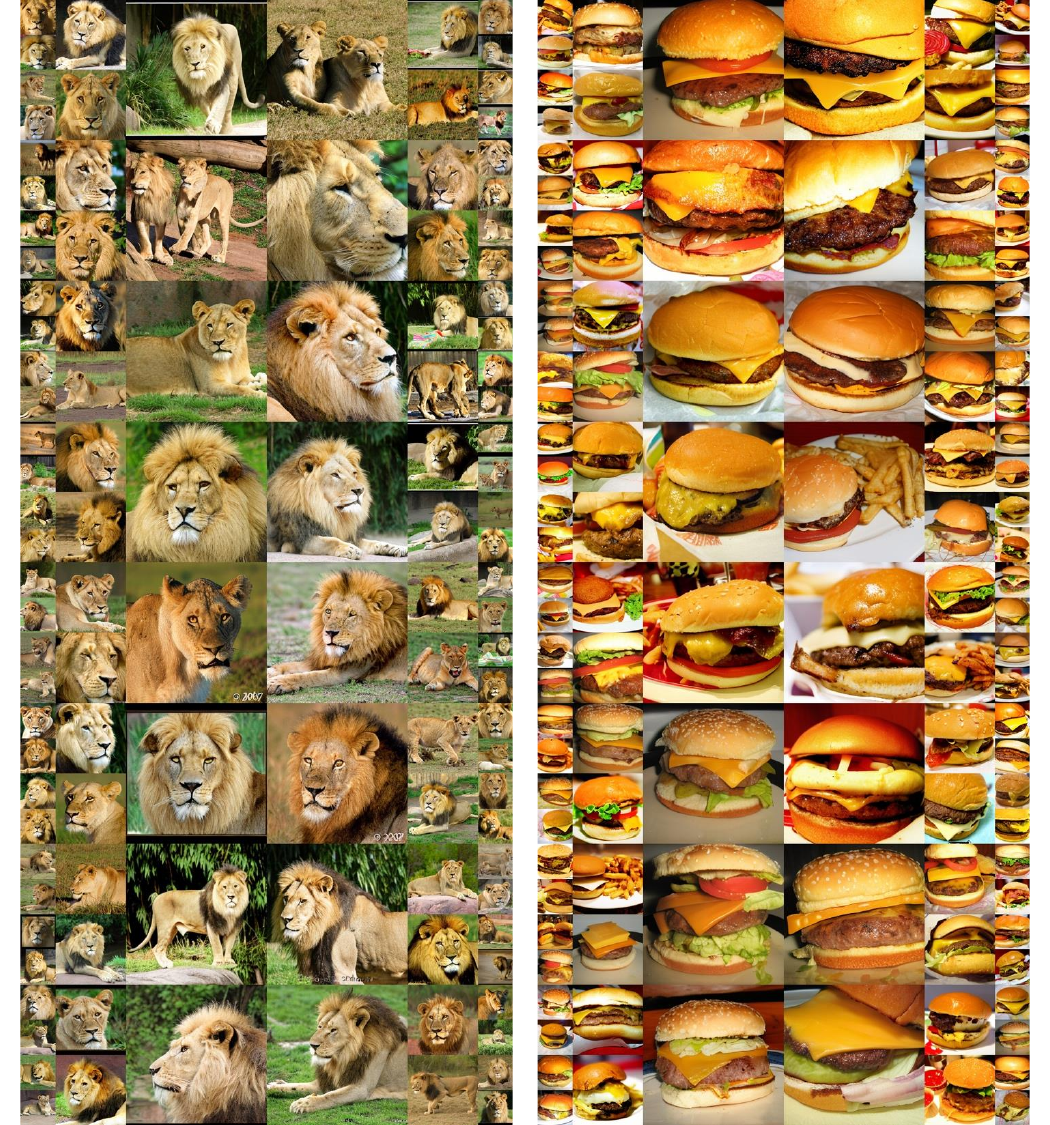}
\vspace{1mm}
\centering
\caption{
  \begin{tabular}{@{}l@{}}
\textbf{Uncurated} 256×256 \textbf{\modelname-XL/2-3T+SSL align~\cite{yu2024representation} Samples (2).} Classifier-free guidance scale = 4.0. \\
(Left): Class = ``lion" (291) \\
(Right): Class = ``cheeseburger" (933)
\end{tabular}
}
\label{fig:qc5}
\end{center}
\end{figure*}

\begin{figure*}[t]
\begin{center}
\includegraphics[width=1.01\linewidth]{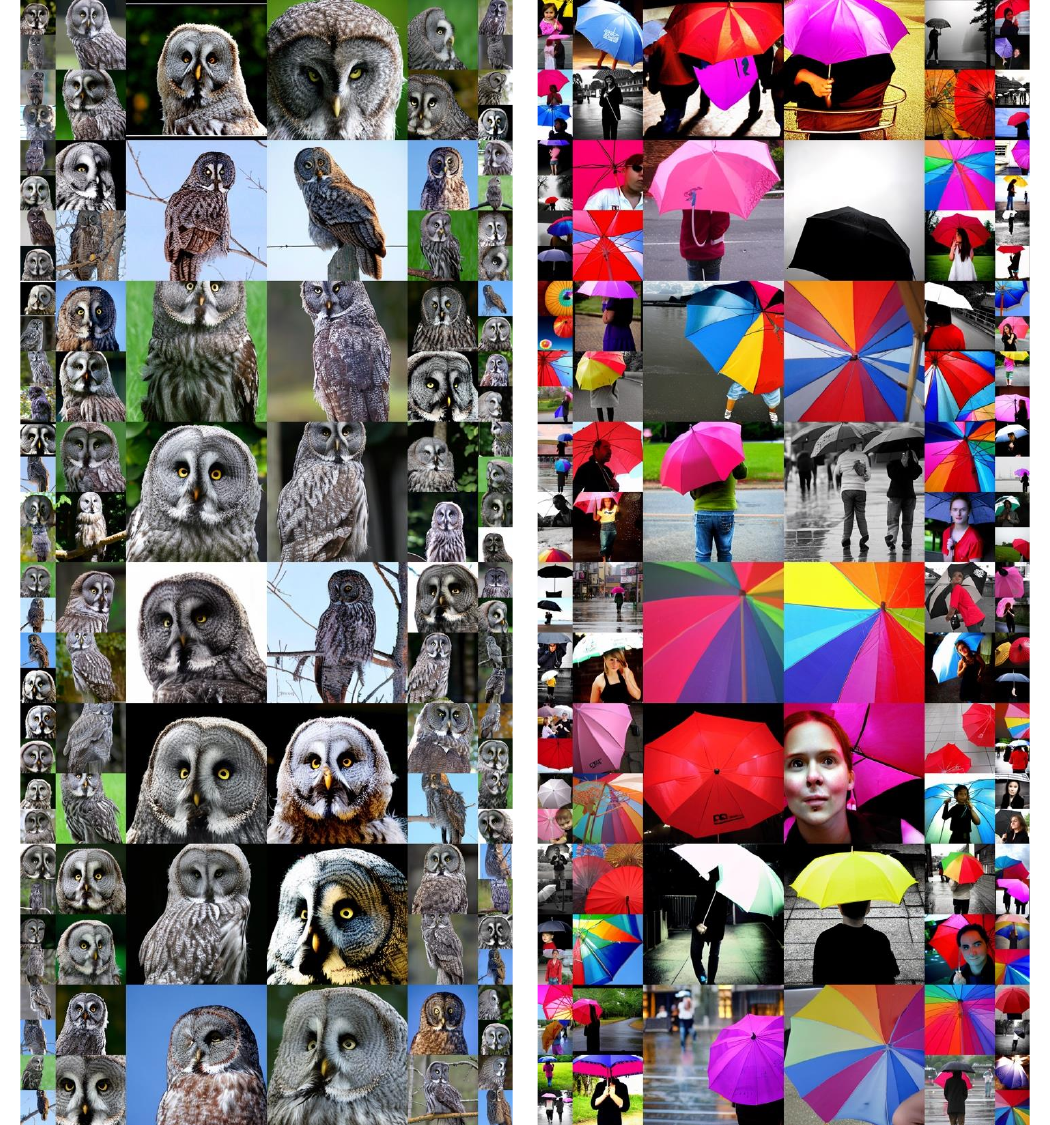}
\vspace{1mm}
\centering
\caption{
  \begin{tabular}{@{}l@{}}
\textbf{Uncurated} 256×256 \textbf{\modelname-XL/2-3T+SSL align~\cite{yu2024representation} Samples (3).} Classifier-free guidance scale = 4.0. \\
(Left): Class = ``great grey owl" (24) \\
(Right): Class = ``umbrella" (879)
\end{tabular}
}
\label{fig:qc6}
\end{center}
\end{figure*}

\begin{figure*}[t]
\begin{center}
\includegraphics[width=0.95\linewidth]{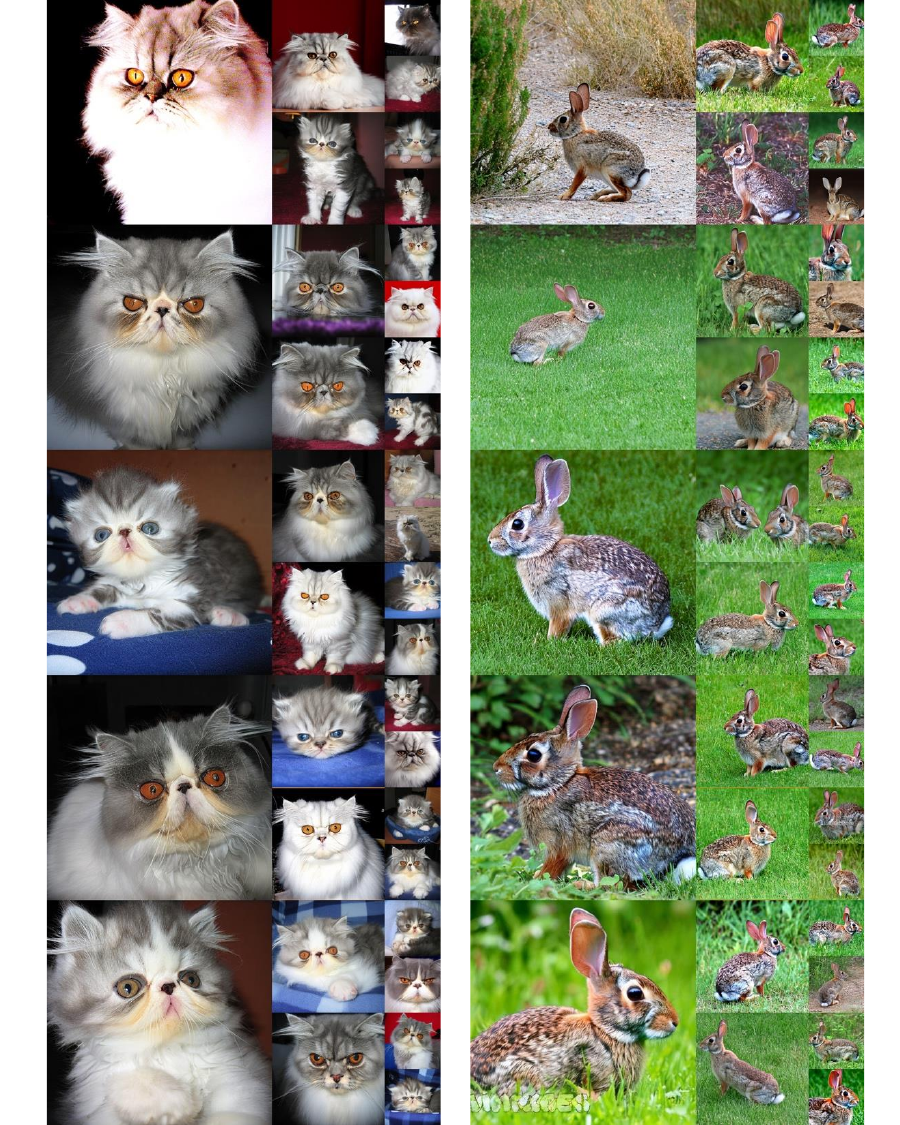}
\vspace{2mm}
\centering
\caption{
  \begin{tabular}{@{}l@{}}
\textbf{Uncurated} 512×512 \textbf{\modelname-XL/2-3T Samples (1).} Classifier-free guidance scale = 4.0. \\
(Left): Class = ``Persian cat" (283) \\
(Right): Class = ``wood rabbit" (330)
\end{tabular}
}
\label{fig:qc7}
\end{center}
\end{figure*}

\begin{figure*}[t]
\begin{center}
\includegraphics[width=0.95\linewidth]{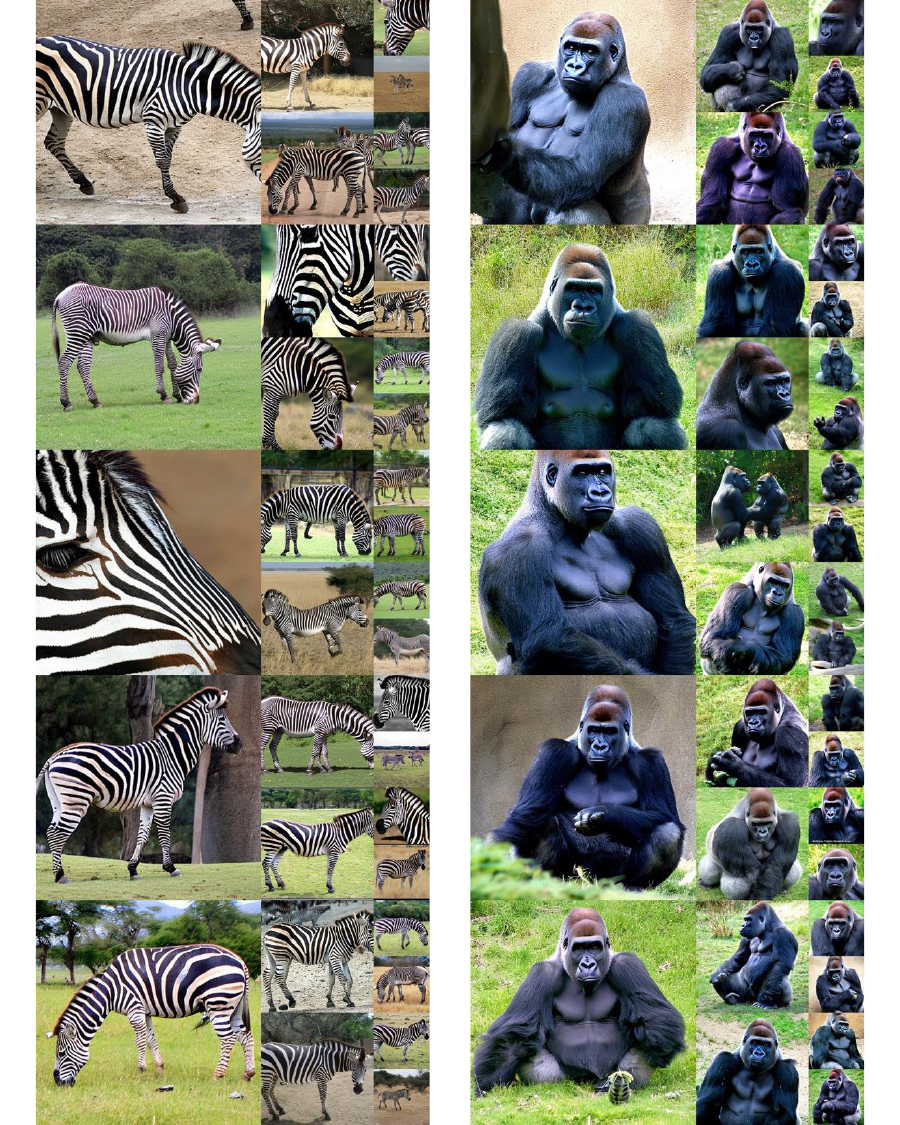}
\vspace{2mm}
\centering
\caption{
  \begin{tabular}{@{}l@{}}
\textbf{Uncurated} 512×512 \textbf{\modelname-XL/2-3T Samples (2).} Classifier-free guidance scale = 4.0. \\
(Left): Class = ``zebra" (340) \\
(Right): Class = ``gorilla" (366)
\end{tabular}
}
\label{fig:qc8}
\end{center}
\end{figure*}

\begin{figure*}[t]
\begin{center}
\includegraphics[width=0.95\linewidth]{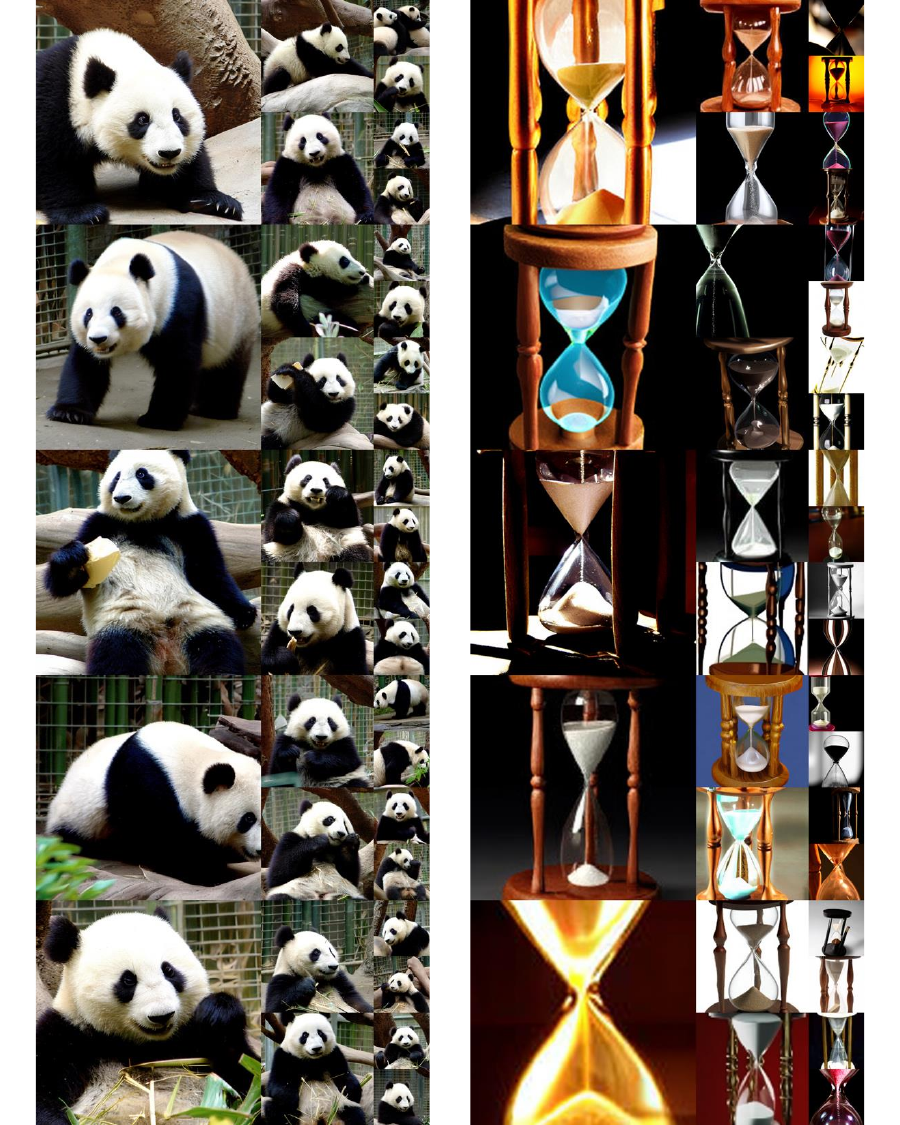}
\vspace{2mm}
\centering
\caption{
  \begin{tabular}{@{}l@{}}
\textbf{Uncurated} 512×512 \textbf{\modelname-XL/2-3T+SSL align~\cite{yu2024representation} Samples (1).} Classifier-free guidance scale = 4.0. \\
(Left): Class = ``giant panda" (388) \\
(Right): Class = ``hourglass" (604)
\end{tabular}
}
\label{fig:qc9}
\end{center}
\end{figure*}

\begin{figure*}[t]
\begin{center}
\includegraphics[width=0.95\linewidth]{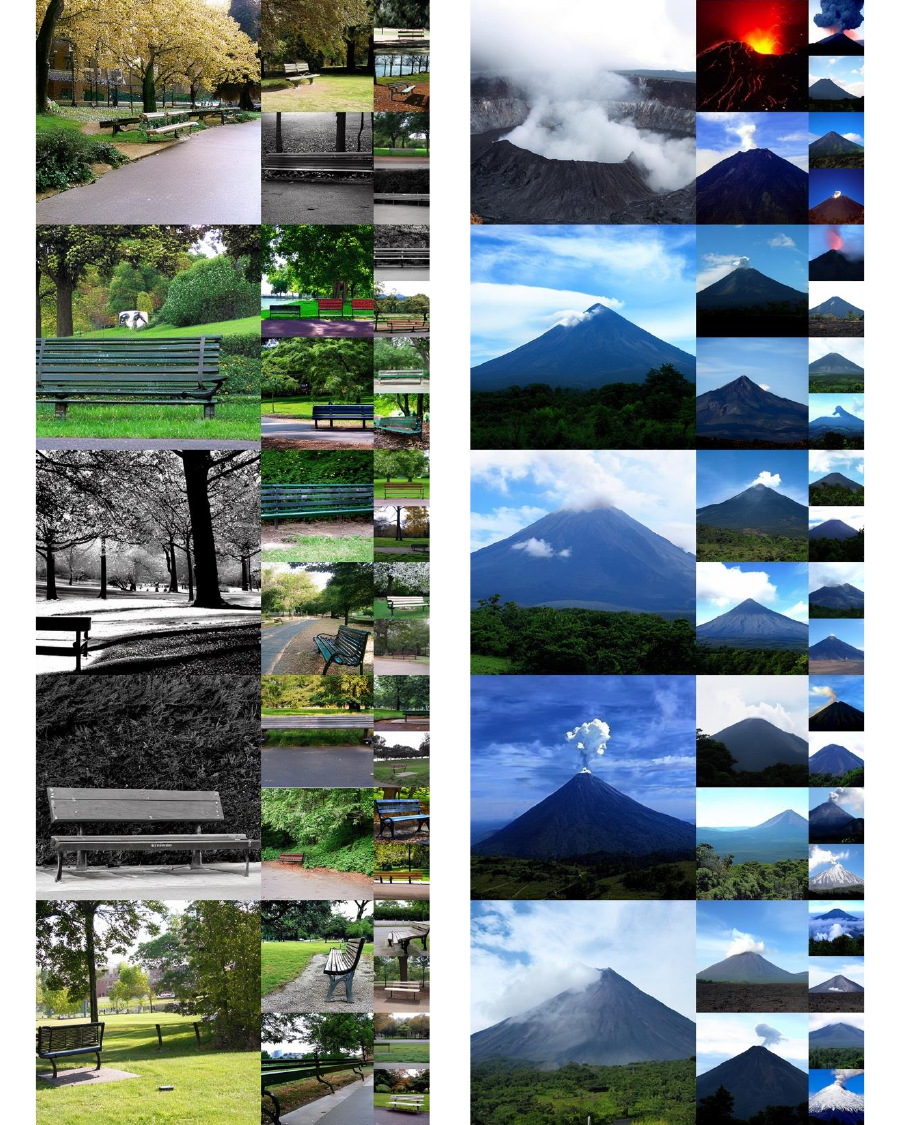}
\vspace{2mm}
\centering
\caption{
  \begin{tabular}{@{}l@{}}
\textbf{Uncurated} 512×512 \textbf{\modelname-XL/2-3T+SSL align~\cite{yu2024representation} Samples (2).} Classifier-free guidance scale = 4.0. \\
(Left): Class = ``park bench" (703) \\
(Right): Class = ``volcano" (980)
\end{tabular}
}
\label{fig:qc10}
\end{center}
\end{figure*}

\begin{figure*}[t]
\begin{center}
\includegraphics[width=1.0\linewidth]{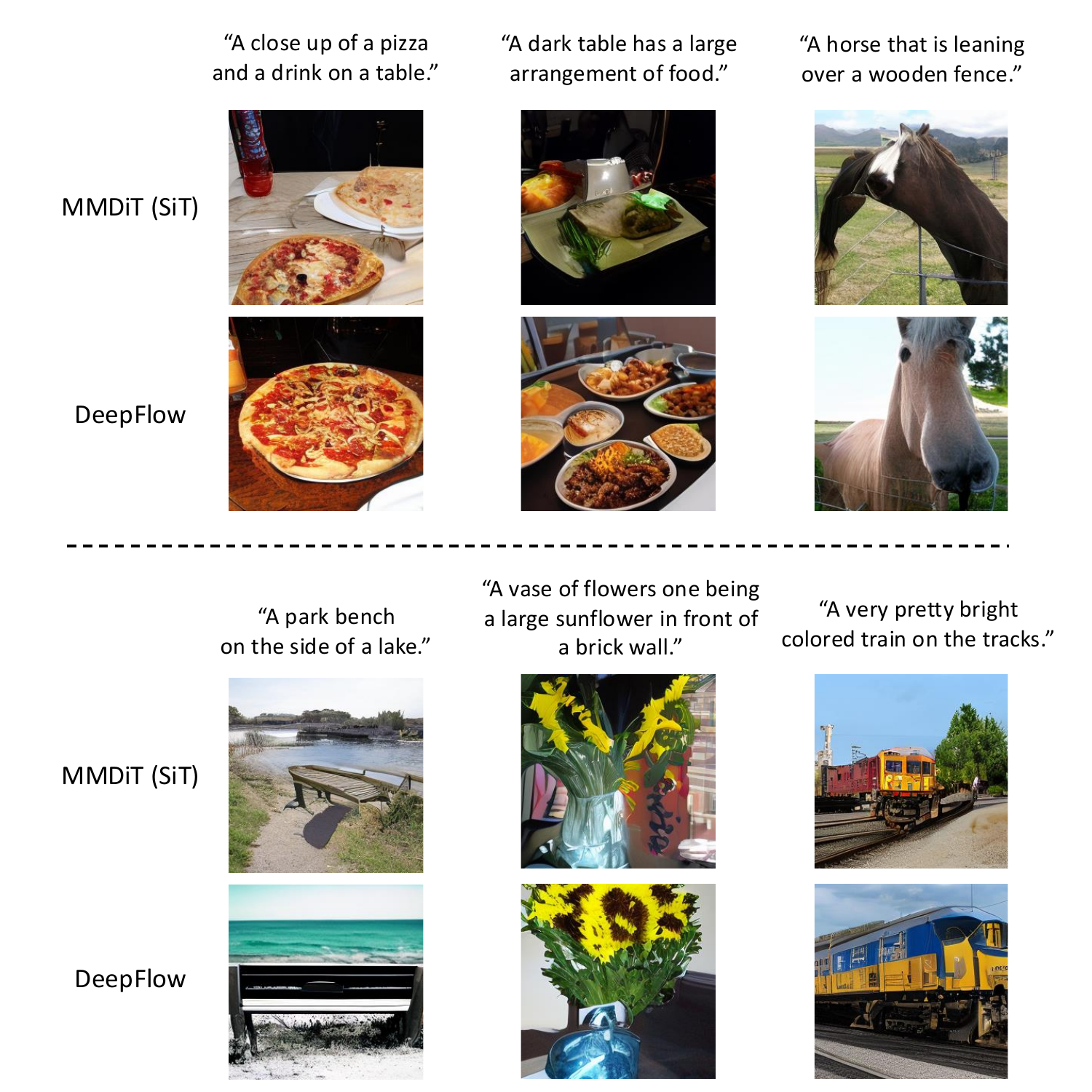}
\vspace{2mm}
\centering
\caption{\textbf{Text-to-Image Generation Results.}
}
\label{fig:qc11}
\end{center}
\end{figure*}

In this section, we provide an extensive qualitative analysis guided by the following criterion: (i) \textit{Supplementary to Figure 4 in the main paper: Can \modelname generate high-quality samples even when trained for substantially fewer epochs?} This question is addressed in \cref{fig:quality_epoch,fig:quality_epoch2}, which visualize samples generated by SiT-XL/2~\cite{ma2024sit} and \modelname-XL/2-3T trained across varying epochs. We observe that \modelname-XL/2-3T not only yields highly promising results at just 80 epochs but also demonstrates stable convergence in subsequent epochs.
(ii) \textit{Can \modelname further enhance its generation capability by leveraging classifier-free guidance (CFG)~\cite{ho2022classifier}?} We demonstrate the visual effectiveness of \modelname-XL/2-3T with CFG by sampling 256×256 images at a CFG scale of 4.0, as illustrated in \cref{fig:qc1,fig:qc2,fig:qc3}. Moreover, we show that the generative performance can be further improved by integrating SSL alignment~\cite{yu2024representation}, as shown in \cref{fig:qc4,fig:qc5,fig:qc6}. Finally, \modelname-XL/2-3T successfully synthesizes high-resolution images (512×512) of superior quality, as demonstrated in \cref{fig:qc7,fig:qc8,fig:qc9,fig:qc10}. (iii) \textit{Can \modelname achieve superior text-to-image generation quality compared to SiT~\cite{ma2024sit}?} \cref{fig:qc11} visually compares samples generated by MMDiT~\cite{esser2024scaling} trained with the SiT objective against those produced by \modelname, using identical text prompts. Notably, \modelname generates more realistic images that also exhibit higher fidelity to the provided textual descriptions.

\subsection{Datasets and Metrics}
\label{sec:dataset}

The datasets we used for training and evaluating \modelname are described as follows:

\vspace{2mm}

\noindent\textbf{ImageNet-1K}: We train and evaluate \modelname on ImageNet-1K dataset for class-conditional generation benchmark. This dataset
spans 1000 object classes and contains 1,281,167 training images, 50,000 validation images and
100,000 test images. The generation results are
evaluated with generation FID using pre-computed statistics and scripts from ADM~\cite{dhariwal2021diffusion}.

\vspace{1mm}
\noindent License: \href{https://image-net.org/accessagreement}{https://image-net.org/accessagreement} 

\vspace{1mm}
\noindent URL: \href{https://www.image-net.org/}{https://www.image-net.org/}

\vspace{2mm}
\noindent\textbf{MS-COCO}: We train and evaluate \modelname on MS-COCO dataset for text-to-image generation benchmark. This dataset contains 82,783 images for training, 40,504 images for validation. 
The generation results are
evaluated with generation FID and FD$_{DINO_{v2}}$~\cite{stein2023exposing}. 

\vspace{1mm}
\noindent License: \href{https://cocodataset.org/#termsofuse}{https://cocodataset.org/termsofuse} 

\vspace{1mm}
\noindent URL: \href{https://cocodataset.org/#home}{https://cocodataset.org}

\noindent\textbf{GenEval}: Baselines and \modelname trained on MS-COCO for text-to-image generation are further evaluated on GenEval dataset~\cite{ghosh2024geneval}. It consists of 553 prompts with four images generated per prompt. Generated samples are evaluated according to various criteria (e.g., Single object, Two object, Counting, Colors, Position, Color attribute).

\noindent\textbf{FID \textit{vs.} FD$_{DINO_{v2}}$}
We carefully select evaluation metrics tailored to each benchmark. For the ImageNet benchmark, we use the FID score because the inception model employed for FID was pre-trained on ImageNet, making it a suitable measure for this dataset. Conversely, for the MS-COCO benchmark, which has a distribution different from ImageNet, we also report FD$_{DINO_{v2}}$~\cite{stein2023exposing}. This metric leverages a DINO$_{v2}$ model pretrained on a more diverse dataset, ensuring a more appropriate evaluation for MS-COCO dataset.

\subsection{Discussion \& Limitations}
\label{sec:dis_lim}
While the proposed \modelname demonstrates impressive performance and training efficiency in image generation tasks, there remains ample scope for further optimization in future work. First, although our text-to-image results are promising compared to previous flow-based models under fair settings, \modelname still underperforms state-of-the-art models (\eg, \cite{kim2025democratizing, esser2024scaling,ren2025beyond}). Training \modelname on large-scale datasets could be a fruitful direction to improve its performance. Second, exploring deeper theoretical insights into \modelname would provide a more thorough validation of our approach. We anticipate that our \modelname will serve as a general framework for flow-based generative model with this further improvement. 

\captionsetup[figure]{justification=centering}

\input{figures/quality_epochs}

%% file: figures/layer_feature_dist.tex
\begin{figure}[t]
\begin{center}
\includegraphics[width=1.04\linewidth]{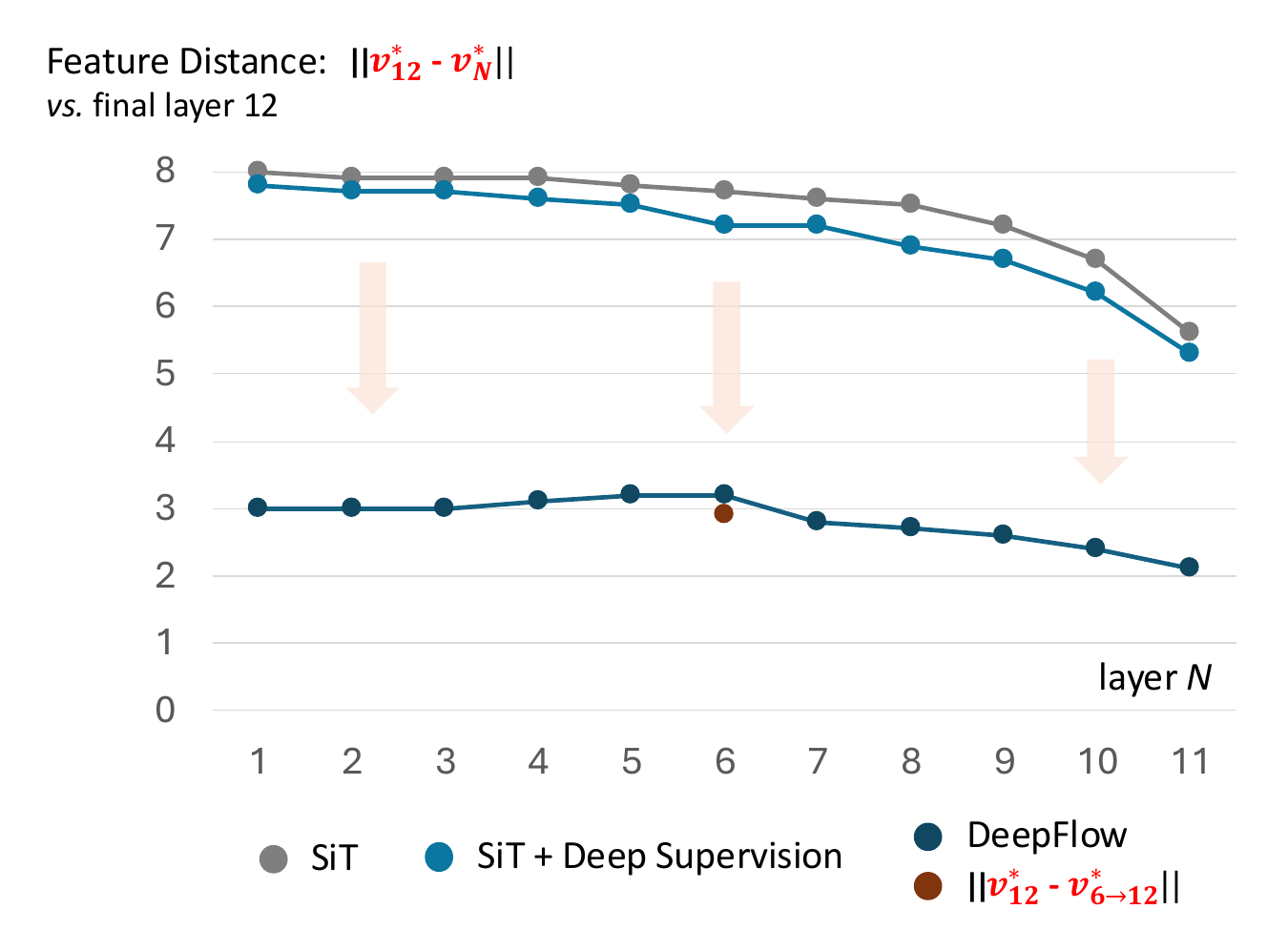}
\caption{
\textbf{Velocity Feature Distance between All Layers and Final Layer.} We provide additional analysis on feature distance to quantify the alignment between velocity features at each layer and one in the final layer. The results demonstrate that \modelname effectively aligns all intermediate features with the final one, even when deep supervision and the \blockname block are applied to a key layer (6th).
}
\label{fig:feature_layer}
\end{center}
\end{figure}

%% file: figures/mit_design.tex
\begin{figure*}[t]
\begin{center}
\includegraphics[width=1.0\linewidth]{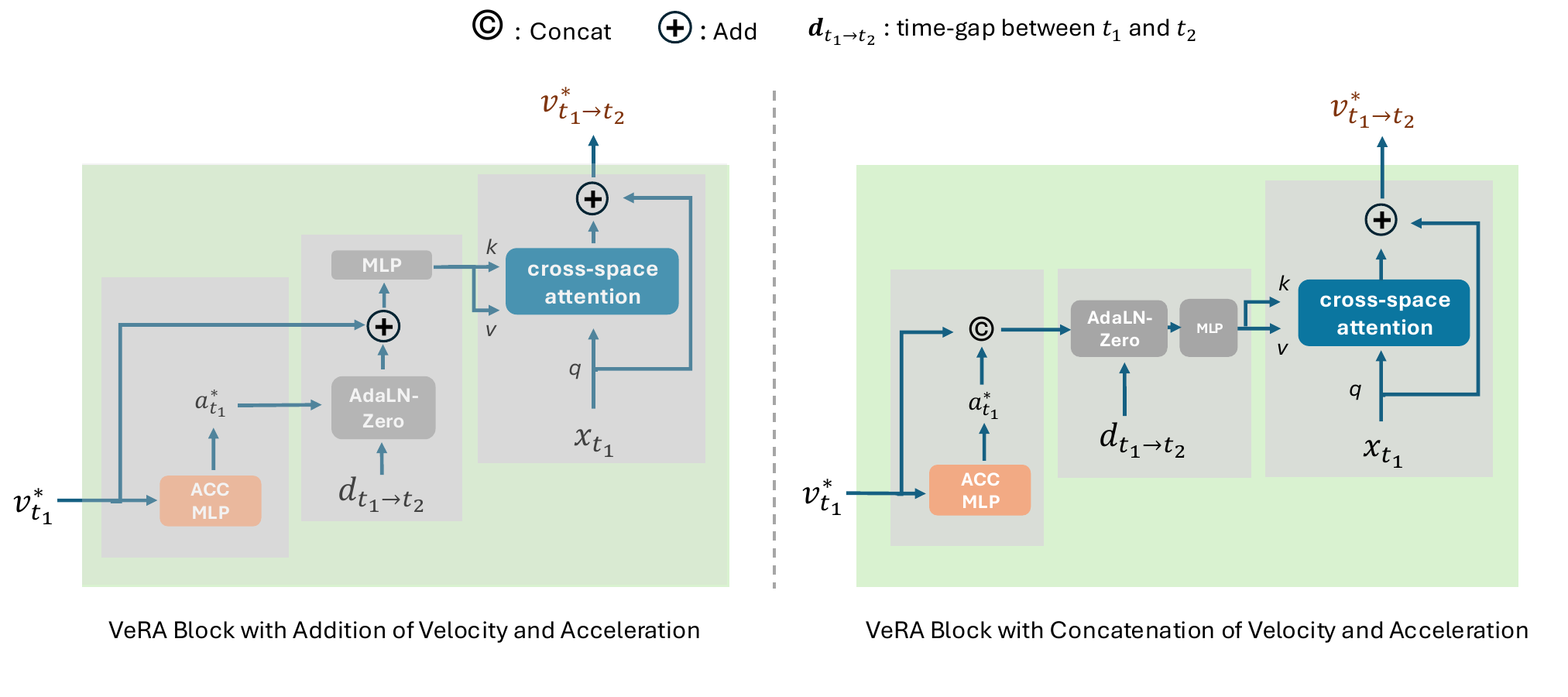}
\caption{
\textbf{Design Choice for \blockname block.} The left panel utilizes addition of velocity and acceleration, while right panel (proposed \blockname block) is differentiated by modulating concatenated feature of velocity and acceleration.
}
\label{fig:mitdesign}
\end{center}
\end{figure*}

%% file: tables/t2i_geneval_v2.tex
\begin{table*}[t]
\centering
\scalebox{0.9}{
\begin{tabular}{
    c|    
    c    
    | c    
    | c    
    c    
    c    
    c    
    c    
    c    
}
 model  & SSL align & Overall$\uparrow$ & Single object & Two object & Counting & Colors & Position & Color attr.  \\
\shline
\multirow{2}{*}{SiT-24~\cite{ma2024sit}}  & \ding{55}   & 0.2672  & 0.8312  & 0.1364  & 0.2062  & 0.4069 & 0.0200  & 0.0025        \\
& DINOv2   & 0.3166  & 0.8969  & 0.2778  & 0.2031 & 0.4495 & 0.0475 & 0.0250  \\
\midrule
\multirow{2}{*}{\modelname-24-3T} & \ding{55}   & 0.2957   & 0.8625  & 0.1919  & 0.2156  & 0.4468 & 0.0250 & 0.0325     \\
& DINOv2    &  0.3458 & 0.9500 & 0.3460  & 0.2406  & 0.4681 & 0.0325 & 0.0375    \\
\end{tabular}}
    \vspace{1mm}
\caption{\textbf{Zero-Shot Text-to-Image Generation Results on GenEval benchmark.} We trained models with MS-COCO~\cite{lin2014microsoft}, following the training setting of REPA~\cite{yu2024representation} and evaluated them with GenEval~\cite{ghosh2024geneval} benchmark.
}
\label{tab:t2i_geneval}
\end{table*}

%% file: tables/cfg_eval.tex
\begin{table}[t]
\centering
\scalebox{0.70}{
\begin{tabular}{
    c|    
    c|    
    c    
    | c    
    c
    c
}
model  & SSL align & CFG & FID$\downarrow$ & sFID$\downarrow$ & IS$\uparrow$ \\
\shline
\multirow{6}{*}{\modelname-XL/2-3T} &  \ding{55}  & 1.3 & 1.98 & 4.39 & 256.7      \\
&   \ding{55}  & \baseline{1.325} & \baseline{1.97} & \baseline{4.39} & \baseline{264.7} \\
&   \ding{55}  & 1.35 & 2.00 & 4.4 & 271.6   \\
\cline{2-6}
&   DINOv2  & 1.275 & 1.78 & 4.45 & 263.4 \\
&   DINOv2   & \baseline{1.3} & \baseline{1.77} & \baseline{4.44} & \baseline{271.3}  \\
 &   DINOv2  & 1.325 & 1.80 & 4.44 & 277.7 \\
\end{tabular}}
\vspace{2mm}
\caption{\textbf{Optimal CFG~\cite{ho2022classifier} Scale Search.} We tested \modelname-XL/2-3T (trained with 400 epochs) with different CFG (classifier-free guidance) scales.
}
\label{tab:cfg_eval}
\end{table}

%% file: figures/sample_steps.tex
\begin{figure}[t]
\begin{center}
\includegraphics[width=1.0\linewidth]{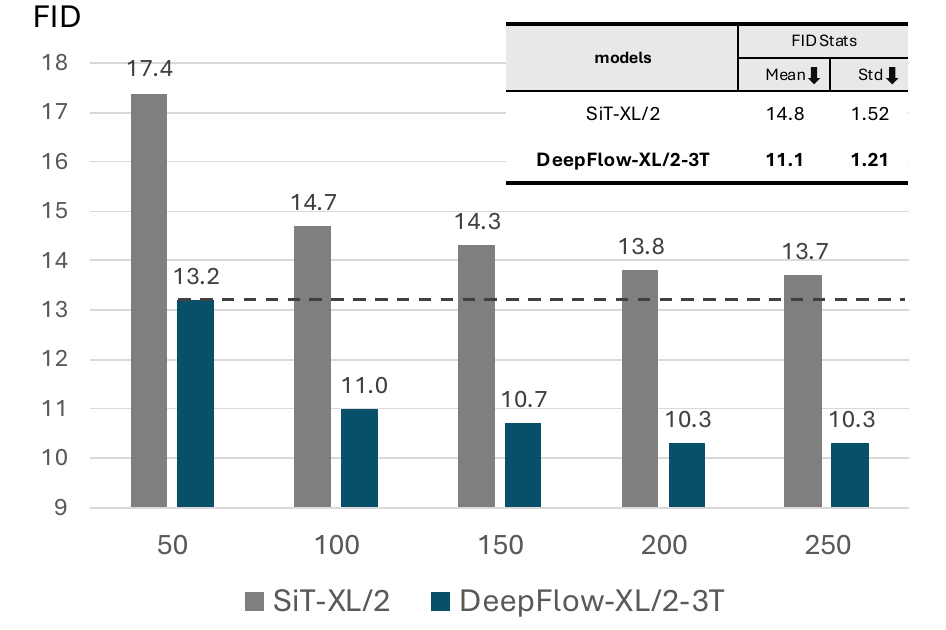}
\caption{
\textbf{Ablation Study on The Number of Sampling Steps.} We provide additional analysis on performance sensitivity of our \modelname-XL/2-3T and SiT-XL/2 to different number of sampling steps including 250, 100, 150, 100, 50 SDE steps.
}
\label{fig:sampling_steps}
\end{center}
\end{figure}

%% file: figures/quality_epochs.tex
\begin{figure*}[t]
\begin{center}
\includegraphics[width=1.0\linewidth]{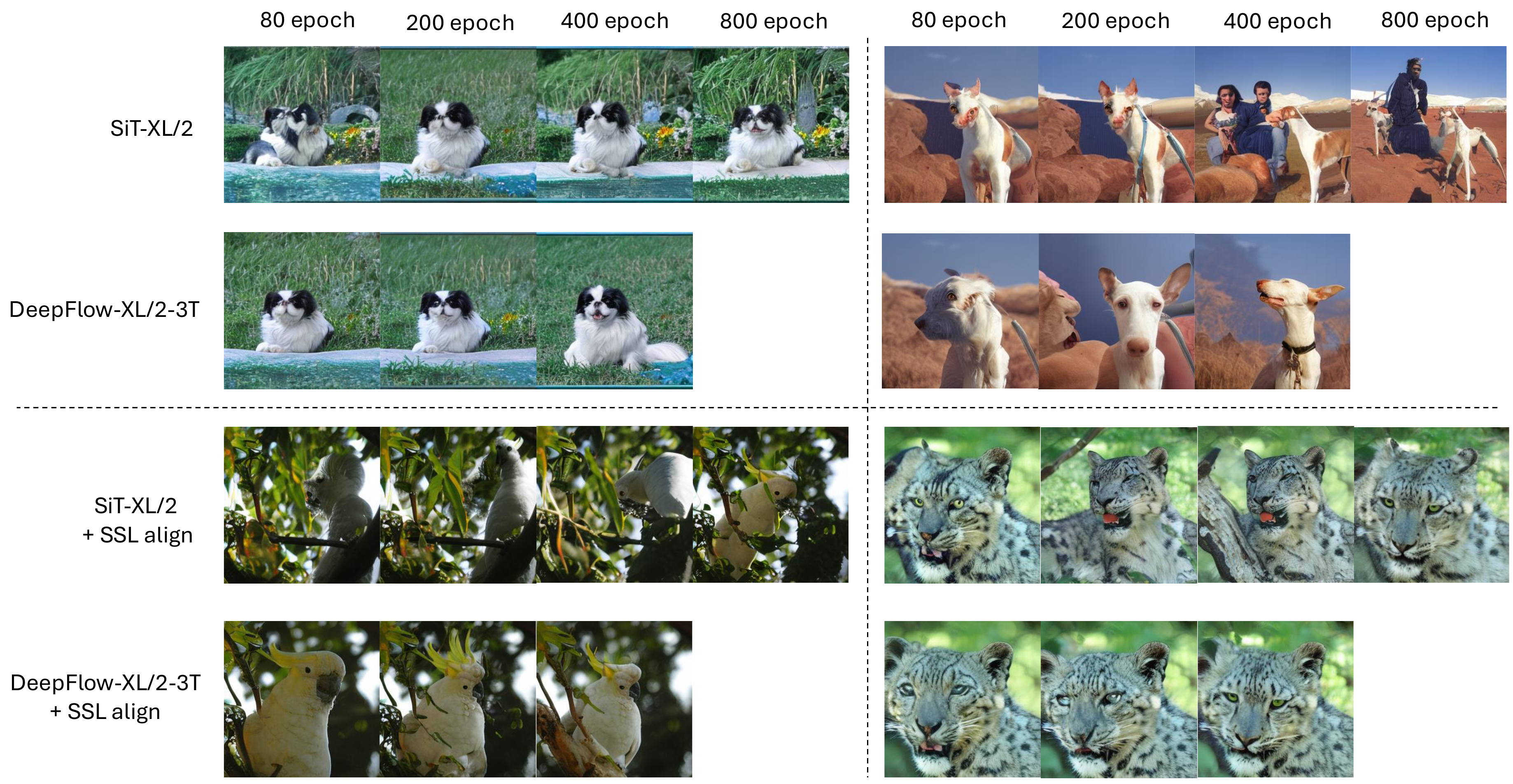}
\caption{
\textbf{Qualitative Comparisons with Baseline in Different Epochs (1).}
}
\label{fig:quality_epoch}
\end{center}
\end{figure*}

\begin{figure*}[t]
\begin{center}
\includegraphics[width=1.0\linewidth]{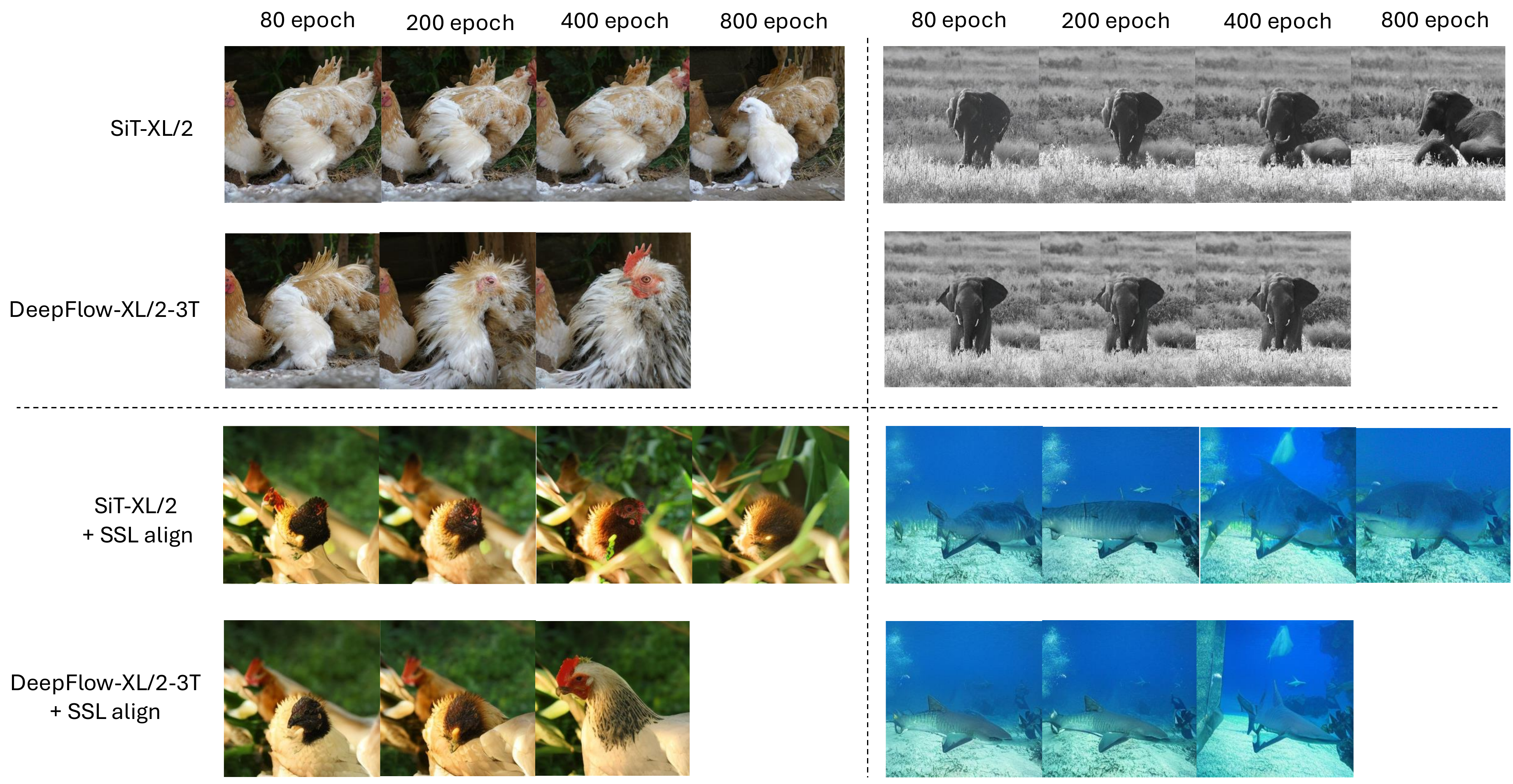}
\caption{
\textbf{Qualitative Comparisons with Baseline in Different Epochs (2).}
}
\label{fig:quality_epoch2}
\end{center}
\end{figure*}